\documentstyle[fleqn]{article}
\begin{document}
\bibliographystyle{plain}
\hyphenation{mono-tony Mono-tony mono-tonic}
\newtheorem{theorem}{Theorem}
\newtheorem{corollary}{Corollary}
\newtheorem{lemma}{Lemma}
\newtheorem{exercise}{Exercise}
\newtheorem{claim}{Claim}
\newtheorem{definition}{Definition}
\newtheorem{example}{Example}
\newtheorem{thesis}{Thesis}
\newenvironment{notation}{\noindent\bf Notation:\em\penalty1000}{}
\newcommand{\blackslug}{\mbox{\hskip 1pt \vrule width 4pt height 8pt 
depth 1.5pt \hskip 1pt}}
\newcommand{\QED}{\quad\blackslug\lower 8.5pt\null\par}
\newcommand{\proof}{\par\penalty-1000\vskip .5 pt\noindent{\bf Proof\/: }}
\newcommand{\ru}{\rule[-0.4mm]{.1mm}{3mm}}
\newcommand{\nni}{\ru\hspace{-3.5pt}}
\newcommand{\sni}{\ru\hspace{-1pt}}
\newcommand{\pre}{\hspace{0.28em}}
\newcommand{\post}{\hspace{0.1em}}
\newcommand{\NIm}{\pre\nni\sim}
\newcommand{\NI}{\mbox{$\: \nni\sim$ }}
\newcommand{\notNIm}{\pre\nni\not\sim}
\newcommand{\notNI}{\mbox{ $\nni\not\sim$ }}
\newcommand{\NIVm}{\pre\nni\sim_V}
\newcommand{\NIV}{\mbox{ $\nni\sim_V$ }}
\newcommand{\notNIVm}{\pre\sni{\not\sim}_V\post}
\newcommand{\notNIV}{\mbox{ $\sni{\not\sim}_V$ }}
\newcommand{\NIWm}{\pre\nni\sim_W}
\newcommand{\NIW}{\mbox{ $\nni\sim_W$ }}
\newcommand{\NIWp}{\mbox{ $\nni\sim_{W'}$ }}
\newcommand{\notNIWm}{\pre\sni{\not\sim}_W\post}
\newcommand{\notNIW}{\mbox{ $\sni{\not\sim}_W$ }}
\newcommand{\eem}{\hspace{0.8mm}\rule[-1mm]{.1mm}{4mm}\hspace{-4pt}}
\newcommand{\EM}{\eem\equiv}
\newcommand{\notEM}{\eem\not\equiv}
\newcommand{\R}{\cal R}
\newcommand{\notR}{\not {\hspace{-1.5mm}{\cal R}}}
\newcommand{\bK}{{\bf K}}
\newcommand{\bKp}{${\bf K}^p$}
\newcommand{\oK}{$\overline {\bf K}$}
\newcommand{\bP}{{\bf P}}
\newcommand{\ga}{\mbox{$\alpha$}}
\newcommand{\gb}{\mbox{$\beta$}}
\newcommand{\gc}{\mbox{$\gamma$}}
\newcommand{\gd}{\mbox{$\delta$}}
\newcommand{\gep}{\mbox{$\varepsilon$}}
\newcommand{\gf}{\mbox{$\zeta$}}
\newcommand{\cA}{\mbox{${\cal A}$}}
\newcommand{\cB}{\mbox{${\cal B}$}}
\newcommand{\cC}{\mbox{${\cal C}$}}
\newcommand{\cE}{\mbox{${\cal E}$}}
\newcommand{\cF}{\mbox{${\cal F}$}}
\newcommand{\cK}{\mbox{${\cal K}$}}
\newcommand{\cL}{\mbox{${\cal L}$}}
\newcommand{\cM}{\mbox{${\cal M}$}}
\newcommand{\cU}{\mbox{${\cal U}$}}
\newcommand{\ab}{\mbox{\ga \NI \gb}}
\newcommand{\cd}{\mbox{\gc \NI \gd}}
\newcommand{\ef}{\mbox{\gep \NI \gf}}
\newcommand{\xe}{\mbox{$\xi$ \NI $\eta$}}  
\newcommand{\pht}{\mbox{$\varphi$ \NI $\theta$}}
\newcommand{\rt}{\mbox{$\rho$ \NI $\tau$}}
\newcommand{\Cn}{\mbox{${\cal C}n$}}
\newcommand{\Pf}{\mbox{${\cal P}_{f}$}}
\newcommand{\ra}{\rightarrow}
\newcommand{\Ra}{\Rightarrow}
\newcommand{\eqdef}{\stackrel{\rm def}{=}}
\newcommand{\absv}[1]{\mid #1 \mid}
\newcommand{\vstar}{\mbox{$V\sstar_{\infty}$}}
\newcommand{\sumstar}{\mbox{$\sum\sstar$}}
\newcommand{\tilh}{\mbox{$\tilde{h}$}}
\newcommand{\tilep}{\mbox{$\tilde{\varepsilon}$}}
\newcommand{\tilf}{\mbox{$\tilde{f}$}}
\newcommand{\gahat}{\mbox{$\hat{\ga}$}}
\newcommand{\gafalse}{\mbox{$\ga\NI{\bf false}$}}
\newcommand{\sstar}{^{*}}
\newcommand{\calR}{\mbox{${\cal R}\sstar$}}

\title{What does a conditional knowledge base entail? 
\thanks{This work was 
partially supported by grant 351/89 from the Basic Research
Foundation, Israel Academy of Sciences and Humanities and
by the Jean and Helene Alfassa fund for 
research in Artificial Intelligence.
Its final version was prepared while the first author was visiting
the Laboratoire d'Informatique Th\'{e}orique et de Programmation,
Univ. Paris 6}}
\author{Daniel Lehmann\thanks{Department of Computer Science, 
Hebrew University, Jerusalem 91904 (Israel)}
\and Menachem Magidor\thanks{Department of Mathematics, 
Hebrew University, Jerusalem 91904 (Israel)}
}
\date{}
\maketitle

\begin{abstract}
This paper presents a logical approach to nonmonotonic
reasoning based on the notion of a nonmonotonic consequence relation.
A conditional knowledge base, consisting of a set of
conditional assertions of the type {\bf if} \ldots {\bf then} \ldots,
represents the explicit defeasible knowledge an agent has about the
way the world generally behaves.  We look for a plausible definition
of the set of all conditional
assertions entailed by a conditional knowledge base.
In a previous paper~\cite{KLMAI:89}, S. Kraus and the authors defined and 
studied {\em preferential} consequence relations.
They noticed that not all preferential relations could 
be considered as reasonable inference procedures.
This  paper studies a more restricted
class of consequence relations, {\em rational} relations.
It is argued that any reasonable nonmonotonic inference procedure
should define a rational relation.
It is shown that the rational relations are exactly those that may 
be represented by a {\em ranked} preferential model, or by a 
(non-standard) probabilistic model.
The rational closure of a conditional knowledge base is defined
and shown to provide an attractive answer to the question of the title.
Global properties of this closure operation are proved: it is a cumulative
operation.
It is also computationally tractable. 
This paper assumes the underlying language is propositional.
\end{abstract}

\section{Introduction}
\label{sec:intro}

\subsection{Background}
\label{subsec:back}
Inference is the process of achieving explicit information that was
only implicit in the agent's knowledge. Human beings are astoundingly
good at infering useful and very often reliable information from
knowledge that seems mostly irrelevant, sometimes erroneous and even
self-contradictory.  They are even better at correcting inferences
they learn to be in contradiction with reality. It is already a
decade now that Artificial Intelligence has realized that the analysis of 
models of such inferences was a major task.

Many nonmonotonic systems have been proposed as formal
models of this kind of inferences. The best known are probably:
circumscription
\cite{McCarthy:80}, the modal systems of \cite{McDer:80} and 
\cite{Moo:85},
default logic \cite{Reiter:80} and negation as failure
\cite{Clark:78}.  An up-to-date survey of the field of nonmonotonic
reasoning may be found in \cite {Reiter:87}.
Though each of these systems is interesting per se, it is not clear
that any one of them really captures the whole generality of
nonmonotonic reasoning.  Recently (see in particular
the panel discussion of \cite{TARAK:88}) 
a number of researchers expressed their disappointment at existing systems and
suggested that no purely logical analysis could be satisfactory.

This work tries to contradict this pessimistic outlook. 
It takes a purely logical
approach, grounded in A.~Tarski's framework of  
consequence relations \cite{Tar:56} and studies the very general notion of 
a {\em sensible} conclusion. It seems that this is a common ground that can be
widely accepted: 
all reasonable inference systems draw only sensible conclusions.
On the other hand, as will be shown, the notion of a sensible conclusion has 
a non-trivial mathematical theory and many interesting properties are 
shared by all ways of drawing sensible conclusions.

The reader is referred to~\cite{KLMAI:89} for a full description of
background, motivation and the relationship of the present approach
with previous work in Conditional Logic. 
We only wish to add here that, even though the present work will be 
compared explicitly only with previous work of E.~Adams, some of the
intuitions developed here are related with intuitions exposed already
in the first works on Conditional Logic, such as~\cite{Ramsey:25} 
or~\cite{Chisholm:46}. The interested reader may find many relevant 
articles in~\cite{IFS:81} and should in particular look at~\cite{Harper:81}.
The main difference between our approach and Conditional Logic is that we take
the view that the truth of a conditional assertion is {\em necessary},
i.e., does not depend on the state of the world. For us, worlds give truth 
values to propositions but not to assertions, preferential models give truth 
values to assertions, but not to propositions. The models we propose are
therefore much simpler than those previously proposed in Conditional Logic
and it is doubtful whether they can shed light on the very complex questions
of interest to the Conditional Logic community.

Notations and terminology conform with
those of~\cite{KLMAI:89}, but the present paper is essentially
self-contained. Preliminary versions of part of the material contained
in this paper appeared in~\cite{LMTR:88,Leh:89}.
In~\cite{KLMAI:89} it was suggested that items of {\em default}, i.e., 
{\em defeasible} information should be represented as {\em conditional
assertions}, i.e., pairs of formulas. 
For example, the information that {\em birds normally fly} 
will be represented by the conditional assertion
\mbox{$b$ \NI $f$}, where $b$ and $f$ are propositional variables
representing {\em being a bird} and {\em flying} respectively.
A set (finite or infinite) of conditional assertions is called a conditional 
knowledge base (knowledge base, in short) and represents the defeasible
knowledge an agent may have. The fundamental question studied in this paper
is the following: given a knowledge base \bK, what are the
conditional assertions that should be considered as entailed, 
i.e., logically implied, by \bK? 
We consider that an assertion \ab\ should be entailed by \bK\ if,
on the basis of the defeasible information contained in \bK\ and knowing
that the proposition \ga\ is true, it would be sensible to conclude
(defeasibly) that \gb\ is true.

The question asked in the title and detailed just above has no simple
answer and has probably no unique answer good for everyone in every 
situation.
It may well be the case that, in different situations or for different domains
of knowledge, the pragmatically {\em right} answers to the question of
the title differ.
This feeling has been recently expressed in~\cite{DoyleWell:91}.
The first part of this paper defines the notion of a {\em rational} set of 
assertions and
defends the thesis that any reasonable answer to the question of the title
must consist of such a set of assertions.
\begin{thesis}
\label{rational}
The set of assertions entailed by any set of assertions is rational.
\end{thesis}
The second part of the paper describes a specific construction, {\em rational 
closure}, and shows that the rational closure of a set of assertions is 
rational.
This construction is then studied and its value as an answer to the question 
of the title assessed. 
We think that, in many situations, this is an acceptable answer, but do not 
claim that it provides an answer suitable to any situation.
We have just argued that such an answer probably does not exist.
One of the main interests of the rational closure construction
is that it provides a proof of the existence of some uniform, well-behaved 
and elegant way of answering the question.
In doing so, we develop criteria by which to judge possible answers.
We shall in particular consider properties of the mapping
from \bK\ to the set of all the assertions it entails and prove that
our construction of the rational closure satisfies them. This effort 
and these results have to be compared with the essential absence,
for the moment, of similar results about the systems of 
nonmonotonic reasoning mentioned above.

\subsection{Plan of this paper}
\label{subsec:plan}
We survey here the main parts of this paper. 
The introductions to the different sections contain a detailed description.
Section~\ref{sec:pref} is devoted to preferential consequence relations.
This family of relations was defined and studied in~\cite{KLMAI:89}.
The first part of this section mainly recalls definitions and results
of~\cite{KLMAI:89}, its last part presents deep new technical
results on preferential entailment that will be used in the sequel,
but it may be skipped on a first reading.
Section~\ref{sec:rat} presents the restricted family of relations that is of
interest to us: rational relations.
This family was first defined, but not studied, 
in~\cite[Section 5.4]{KLMAI:89}.
The main result of this section is a representation theorem characterizing
rational relations in terms of ranked models.
Section~\ref{sec:entrank} shows that entailment with respect to ranked models
is exactly entailment with respect to preferential models and 
provides an alternative
proof of E.~Adams'~\cite{Adams:75} characterization of preferential entailment
in terms of his probabilistic semantics.
Appendix~\ref{appen:nonstandard} describes a family of models based on 
non-standard (in the sense of A.~Robinson) probability models and 
shows that these models provide another 
exact representation for rational consequence relations. 
This provides us with a strong justification for considering rational
relations.
Section~\ref{sec:ratclos} draws on all previous sections and is the heart of this paper.
It proposes an answer to the question of the title.
The notion of rational closure is first defined abstractly and 
global properties proved. It is then showed that finite knowledge bases 
have a rational closure and a model-theoretic construction is provided.
An efficient algorithm is proposed for computing the rational closure of a 
finite knowledge base.
We then discuss some examples, remark that rational closure does not provide
for inheritance of generic properties to exceptional classes, and
finally propose a second thesis.

\section{Preferential relations and  models}
\label{sec:pref}
\subsection{Introduction}
\label{subsec:intropref}
The first part of this section, i.e., 
Sections~\ref{subsec:prefrel}--\ref{subsec:prefmod},
recalls definitions and results of~\cite{KLMAI:89} and provides 
an example (new) of a preferential relation that cannot be defined by a 
well-founded model.
Then, in Section~\ref{subsec:prefent}, the definition and some properties
of preferential entailment are recalled from~\cite{KLMAI:89} 
and some new remarks included. 
Preferential entailment is a fundamental notion that is used throughout the 
paper. 
The last three sections are essentially independent of each other.
They present an in-depth study of preferential entailment.
In a first reading, they should probably be read only cursorily.
The results of Section~\ref{subsec:some} expand on part of~\cite{Leh:89}
and are used in Section~\ref{subsec:rankpref}.
Section~\ref{subsec:ranking} presents a new technique to study
preferential entailment (i.e., ranking). 
It is fundamental from Section~\ref{subsec:proof} and onwards. 
Section~\ref{subsec:comppref} shows that preferential entailment 
is in the class co-NP, and hence is an co-NP-complete problem.
A preliminary version of this last result appeared in~\cite{Leh:89}.

\subsection{Preferential relations}
\label{subsec:prefrel}

Our first step must be to define a language in which to express the basic 
propositions. In this paper Propositional Calculus is chosen. 
Let ${\cal L}$ be the set of well formed  propositional formulas 
(thereafter formulas) over a set of propositional variables.
If the set of propositional variables chosen is finite, 
we shall say that ${\cal L}$ is logically finite.
The classical propositional connectives will be denoted by 
$\neg , \vee , \wedge , \rightarrow$ and $\leftrightarrow$.
The connective $\rightarrow$ therefore denotes material
implication.
Small Greek letters will be used to denote formulas.

A world is an assignment of truth values to the propositional variables.
The set ${\cal U}$ is the set of all worlds.
The satisfaction of a formula by a world is defined as usual.
The notions of satisfaction of a set of formulas, validity of a 
formula and 
satisfiability of a set of formulas are defined as usual. 
We shall write $\models \alpha$ if 
$\alpha$ is valid, i.e., iff
$\forall u \in {\cal U}$, $u \models \alpha$.

If $\alpha$ and $\beta$ are formulas then the pair
\mbox{$\alpha \NI \beta$} 
(read ``from $\alpha$ sensibly conclude~$\beta$'')
is called a conditional assertion. A conditional assertion is a syntactic 
object to 
which the reader may attach any meaning he wants, but the meaning we attach
to such an assertion, and against which the reader should check the logical 
systems
to be presented in the upcoming sections, is the following: if 
$\alpha$  represents the information I have about the true state of the world,
I will jump to the conclusion that $\beta$ is true.
A conditional knowledge base is any set of conditional assertions. Typically
it is a finite set, but need not be so.
Conditional knowledge bases seem to provide a terse and versatile way
of specifying defeasible information.
They correspond to the explicit information an agent may have.

Certain well-behaved sets of conditional assertions will be deemed worthy of
being called {\em consequence relations}. 
We shall use the notation usual for binary relations to
describe consequence relations. So, if $\NI$ is a consequence relation,
\mbox{$\alpha \NI \beta$} indicates that the pair 
\mbox{$\langle \alpha , \beta \rangle$} is in the consequence relation
$\NI$ and
\mbox{$\alpha \notNI \beta$} indicates it is not in the relation.
Consequence relations correspond to the implicit
information an intelligent agent may have. 
Consequence relations are typically infinite sets. 

Certain especially interesting properties of sets of conditional assertions
(i.e., binary relations on ${\cal L}$) will be described and discussed now.
They are presented in the form of inference rules.
Consequence relations are expected to satisfy those properties.

\begin {equation}
\label{eq:LLE}
{{\models \alpha \leftrightarrow \beta \ \ , \ \  \alpha  \NI \gamma} \over
    {\beta \NI \gamma}} \hspace {0.55cm} 
{\rm ({\bf Left \ Logical \ Equivalence}) }
\end {equation}
\begin {equation}
{{\models \alpha \ra \beta\ \ , \ \  \gamma \NI \alpha } 
\over {\gamma \NI \beta}} \hspace {0.6cm} {\rm ({\bf Right \ Weakening})}
\end {equation}
\begin {equation}
\alpha \NI \alpha \hspace {2.9cm} {\rm ({\bf Reflexivity})}
\end {equation}
\begin{equation}
{{\alpha \NI \beta \ \ , \ \ \alpha \NI \gamma} \ \ \  \over
{\alpha \NI \beta \wedge \gamma}} \hspace{0.8cm}{\rm ({\bf And})}
 \end{equation}
\begin{equation}
{{\alpha  \NI \gamma \ \ , \ \ \beta  \NI \gamma} \over
{\alpha \vee \beta  \NI \gamma}} \hspace {1.15cm} {\rm ({\bf Or})}
\end{equation}
\begin {equation}
\label{eq:CM}
{{\alpha \NI \beta \ \ , \ \ \alpha \NI \gamma} 
\over {\alpha \wedge \beta
\NI \gamma}}\hspace {1.15cm} {\rm ({\bf Cautious \  Monotonicity})}
\end {equation}

\begin{definition}
\label{def:prefcons}
A set of conditional assertions that satisfies all six properties above is
called a {\em preferential} consequence relation.
\end{definition}

A more leisurely introduction with motivation may be found 
in~\cite{KLMAI:89} where a larger family of consequence relations,
that of cumulative relations, was also studied. 
This family is closely related to the cumulative inference operations 
studied by D.~Makinson in \cite{Mak:89}.
The attentive reader of~\cite{KLMAI:89} may have noticed that, there,
we reserved ourselves an additional degree of freedom, that we have denied
ourselves here. There, we allowed \cU\ to be a subset of the set of all worlds
and considered the $\models$ symbol appearing in 
{\bf Left Logical Equivalence} and in {\bf Right Weakening} to be interpreted 
relatively to this subset.
This was felt necessary to deal with {\em hard constraints}.
In this work, we shall suppose that a hard constraint \ga\ is interpreted as
the {\em soft} constraint, i.e., the assertion,
\mbox{$\neg \alpha$ \NI {\bf false}}, which was recognized as equivalent
to considering \cU\ to be the set of all worlds satisfying \ga\ 
in~\cite[page 174]{KLMAI:89}. The second proposal there, i.e., to consider
\ga\ to be part of the facts, would not be consistent with our treatment
of rational closure.

For the reader's ease of mind we shall mention two important derived rules.
Both {\bf S} and {\bf Cut} are satisfied by any preferential relation.
\begin {equation}
{{\alpha \wedge \beta \NI \gamma} \over
{\alpha \NI \: \beta \rightarrow \gamma}}\hspace {2.2cm} {\rm ({\bf S})}
\end {equation}
\begin {equation}
{{\alpha \wedge \beta \NI \gamma \ \ , \ \ \alpha \NI \beta} \over 
{\alpha \NI \gamma \;}} \hspace {0.8cm} {\rm ({\bf Cut})}
\end {equation}
The rule of Cut is presented here in a form that is not the most usual one.
Notice, in particular, that we require the left-hand side of the second
assumption to be part of the left-hand side of the first assumption.
This version of Cut is close to the original form proposed by G.~Gentzen.
The following form, more usually used now, is {\em not} acceptable since
it implies monotonicity.
\begin{equation}
{{\alpha \wedge \beta \NI \gamma \ \ , \ \ \alpha' \NI \beta} \over
{\alpha \wedge \alpha' \NI \gamma \;}}
\end{equation}

\subsection{Preferential models and representation theorem}
\label{subsec:prefmod}
The following definitions are also taken from~\cite{KLMAI:89} and justified
there.
We shall define a class of models that we call {\em preferential} 
since they represent a slight variation on those
proposed in \cite{Shoham:87}.
The differences are nevertheless technically important.

Preferential models give a model-theoretic
account of the way one performs nonmonotonic inferences. 
The main idea is that the agent has, in his mind, a partial ordering on 
possible states of the world. State
$s$ is less than state $t$, if, in the agent's mind, $s$ is {\em preferred}
to or more {\em natural} than $t$. The agent is willing to conclude
$\beta$ from $\alpha$, if all {\em most natural} states that satisfy $\alpha$
also satisfy $\beta$.

Some technical definitions are needed.
Let $U$ be a set and $\prec$ a strict partial order on $U$, i.e., a binary
relation that is antireflexive and transitive.
\begin{definition}
\label{def:min}
Let \mbox{$V \subseteq U$}.
We shall say that \mbox{$t \in V$} is minimal in $V$ iff there is no
\mbox{$s\in V$}, such that
\mbox{$s \prec t$}.
We shall say that \mbox{$t \in V$} is a minimum of $V$ iff for every
\mbox{$s\in V, s \neq t$}, we have \mbox{$t \prec s$}.
\end{definition}
\begin{definition}
\label{def:smooth}
Let \mbox{$V \subseteq U$}.
We shall say that $V$ is  smooth iff \ 
\mbox{$\forall t \in V$}, either 
\mbox{$\exists s$} minimal in $V$, such that \mbox{$s \prec t$} or
$t$ is itself minimal in $V$.
\end{definition}
We may now define the family of models we are interested in.

\begin{definition}
\label{def:prefmod}
A {\em preferential} model $W$ is a triple 
\mbox{$\langle S , l , \prec \rangle$}
where $S$ is a set, the elements of which will be called states, 
\mbox{$l:S \mapsto {\cal U}$} assigns a world to each state
and $\prec$ is a strict partial order on $S$
satisfying the following 
{\em smoothness condition}:
\mbox{$\forall \alpha \in L$}, the set of states 
\mbox{$\widehat{\alpha} \eqdef
\{ s \mid s \in S, \: s \EM \alpha \}$} is smooth, where \mbox{$\EM$} 
is defined as
\mbox{$s \EM \alpha$} (read $s$ satisfies $\alpha$) iff 
\mbox{$l(s) \models \alpha$}.
The model $W$ will be said to be finite iff $S$ is finite.
It will be said to be well-founded iff \mbox{$\langle S , \prec \rangle$}
is well-founded, i.e., iff there is no infinite descending chain of states.
\end{definition}

The smoothness condition is only a technical condition.
It is satisfied in any well-founded preferential model,
and, in particular, in any finite model.
When the language \cL\ is logically finite, we could have limited
ourselves to finite models and forgotten the smoothness condition. 
Nevertheless, Lemma~\ref{le:well} will show that, in the general case, 
for the representation
result of Theorem~\ref{compth:pref} to hold we could not have required 
preferential models to be well-founded.
The requirement that the relation $\prec$ be a strict partial order has been
introduced only because such models are nicer and the smoothness condition
is easier to check on those models, but the soundness result 
is true for the larger family of models, where $\prec$ is
just any binary relation (Definitions~\ref{def:min} and \ref{def:smooth} also
make sense for any binary relation $\prec$).
In such a case, obviously, the smoothness condition cannot be dropped even
for finite models.
The completeness
result holds, obviously, also for the larger family,
but is less interesting.

We shall now describe the consequence relation defined by a model.
\begin{definition}
\label{def:prefent}
Suppose a model \mbox{$W = \langle S , l, \prec \rangle$}
and \mbox{$\alpha ,\beta\in L$} are given. The consequence relation 
defined by $W$ will be denoted by \mbox{$\NIW$} and is defined by:
\mbox{$\alpha \NIW \beta$} iff for any $s$ minimal in $\widehat{\alpha}$,  
\mbox{$s \EM \beta$}. 
\end{definition}

If \mbox{$\alpha \NIW \beta$} we shall say that the model $W$ satisfies
the conditional assertion 
\mbox{$\alpha \NI \beta$}, or that $W$ is a model of 
\mbox{$\alpha \NI \beta$}.

The following theorem characterizes preferential consequence relations.
\begin{theorem}[Kraus, Lehmann and Magidor]
\label{compth:pref}
A binary relation \mbox{$\NI$} on ${\cal L}$ is a preferential consequence relation 
iff it is
the consequence relation defined by some preferential model.
If the language ${\cal L}$ is logically finite, then every preferential consequence
relation is defined by some finite preferential model.
\end{theorem}

The next result shows we could not have restricted ourselves to 
well-founded models.

\begin{lemma}
\label{le:well}
There is a preferential relation that is defined by no well-founded 
preferential model.
\end{lemma}
\proof
Let ${\cal L}$ be the propositional calculus on the variables 
\mbox{$p_i , i \in \omega$} ($\omega$ is the set of natural numbers).
We shall consider the model \mbox{$W \eqdef \langle V , l , \prec \rangle$}
where $V$ is the set \mbox{$\{s_i \mid i \in \omega \cup \{ \infty \}\}$},
\mbox{$s_i \prec s_j$} iff \mbox{$i > j$} (i.e., there is an
infinite descending chain of states with a bottom element) and
\mbox{$l(s_i)(p_j)$} is {\bf true} iff \mbox{$j \geq i$}, for
\mbox{$i \in \omega \cup \{ \infty \}$} and \mbox{$j \in \omega$}.
The smoothness property is satisfied since the only subsets of 
$V$ that do not have a minimum are infinite sets $A$ that do not
contain $s_\infty$ and any $\alpha \in L$ that is satisfied in all
states of such a set $A$ is also satisfied in $s_\infty$.
The model $W$ defines a preferential relation \NIW such that 
\mbox{$\forall i \in \omega$} , \mbox{$p_i$ \NIW $p_{i + 1}$} and
\mbox{$p_{i + 1}$ \NIW $\neg p_i$}, but \mbox{$p_0$ \notNIW {\bf false}}.
But clearly, any preferential model defining such a relation must contain
an infinite descending chain of states.
\QED

We do not know of any direct characterization of those relations 
that may be defined by well-founded preferential models.
But Lemma~\ref{le:co2} will show that many relations may be 
defined by well-founded preferential models.
It is clear, though, that the canonical preferential model 
provided by the proof of Theorem~\ref{compth:pref} is rarely well-founded.
Consider, for example, the preferential closure of the empty knowledge base
on a logically infinite language ${\cal L}$.
It may be defined by some well-founded preferential model (the order $\prec$
is empty).
But its canonical model is not well-founded (consider states whose second 
components are larger and larger disjunctions).
We may only make the following obvious remark:
if the underlying language ${\cal L}$ is logically finite,
then all canonical models are well-founded.

\subsection{Preferential entailment}
\label{subsec:prefent}

Now that we have a proof-theoretic definition of a class of relations,
a class of models and a representation theorem relating them, it is natural
to put down the following definition. It will serve us as a first approximate
answer to the question of the title.

\begin{definition}
\label{p-implic}
The assertion ${\cal A}$ is {\em preferentially entailed} by \/{\bf K}
iff it is satisfied by all preferential models of \/{\bf K}.
The set of all conditional assertions that are preferentially entailed
by \/{\bf K} will be denoted by \/${\bf K}^p$. The preferential consequence
relation \/${\bf K}^p$ is called the preferential closure of \/{\bf K}.
\end{definition}
In~\cite{KLMAI:89} it was noted that the
characterization of preferential consequence relations obtained in 
Theorem~\ref{compth:pref} enables us to prove the following.

\begin{theorem}
\label{log:imp2}
Let {\bf K} be a set of conditional assertions, and 
${\cal A}$ a conditional assertion.
The following conditions 
are equivalent:
\begin{enumerate}
\item ${\cal A}$ is preferentially entailed by {\bf K}, i.e., 
${\cal A} \in {\bf K}^p$
\item ${\cal A}$  has a proof from {\bf K} in 
the system {\bf P} consisting of the Rules~\ref{eq:LLE} to~\ref{eq:CM}.
\end{enumerate}
\end{theorem}
The following compactness result follows.
\begin{corollary}[compactness]
\label{comp:pref}
{\bf K} preferentially entails ${\cal A}$ 
iff a finite subset of \/{\bf K} does.
\end{corollary}
The following also follows from Theorem~\ref{log:imp2}.
\begin{corollary}
\label{co1}
The set \/${\bf K}^p$, considered as a consequence relation, is
a preferential consequence relation, therefore there is a preferential
model that satisfies exactly the assertions of \/${\bf K}^p$.
If \/{\bf K} is itself a preferential consequence relation then
\/${\bf K} = {\bf K}^p$.
The set \/${\bf K}^p$ grows monotonically with \/{\bf K}.
\end{corollary}
We see that the operation \mbox{$\bK \mapsto {\bf K}^p$}
is a compact monotonic consequence operation in the sense of 
Tarski~\cite{Tar:56}.
We have a particular interest in finite knowledge bases.
It is therefore useful to put down the following definition.
\begin{definition}
\label{def:fingen}
A preferential consequence relation is {\em finitely generated}
iff it is the preferential closure of a finite knowledge base.
\end{definition}
Lemma~\ref{le:co2} will show that finitely generated relations have 
interesting properties.
In~\cite{KLMAI:89}, it was shown that any preferential relation defines
a strict ordering on formulas by:
\mbox{$\alpha < \beta$} iff \mbox{\ga $\vee$\gb \NI \ga}
and \mbox{\ga $\vee$\gb \notNI \gb}.
\begin{definition}
\label{def:wellf}
A preferential relation is {\em well-founded} iff the strict ordering
relation $<$ it defines is well-founded.
\end{definition}
The following is easy to show.
\begin{lemma}
\label{le:wellfcan}
A preferential relation is well-founded iff the canonical model built
in the proof of Theorem~\ref{compth:pref} is well-founded.
\end{lemma}
We noticed, at the end of 
Section~\ref{subsec:prefmod}, that not all preferential relations that
may be defined by well-founded preferential models are well-founded.

\begin{lemma}
\label{le:co2}
Any finitely generated preferential relation is defined by some well-founded
preferential model.
\end{lemma}
\proof
Let \bK\ be any finite set of assertions.
Let $L_i , i \in \omega$ be an infinite sequence of larger and larger
{\em logically finite} sublanguages of $L$ such that every $L_i$ contains
all the formulas appearing in the assertions of \bK\ and such that
$L$ is the union of the $L_i$'s.
By Theorem~\ref{compth:pref}, for each $i$ there is a {\em finite}
preferential model $W'_i$ that defines the preferential closure of \bK\
over $L_i$.
Let $W_i$ be the finite preferential model (over $L$) obtained by
extending the labeling function of $W'_i$ to the variables of $L - L_i$
in some arbitrary way.
Clearly $W_i$ is a preferential model of \bK.
Let $W$ be the structure obtained by putting all the $W_i$'s one
alongside the other (the partial ordering $\prec$ on $W$ never
relates states belonging to $W_i$'s with different $i$'s).
The structure $W$ is well-founded, therefore satisfies the smoothness condition
and is a preferential model.
Any assertion that is preferentially entailed by \bK\ (over $L$) is
satisfied by every $W_i$, and is therefore satisfied by $W$.
For any assertion \cA\ that is not preferentially entailed by
\bK\, one may find a language $L_i$ large enough to include the formulas
of \cA. Over $L_i$, the assertion \cA\ is not preferentially
entailed by \bK, by Theorem~\ref{log:imp2}, since a proof in the 
small language is a proof in the larger one.  
Therefore $W'_i$ does not satisfy \cA.
We conclude that $W_i$ does not satisfy \cA\ and that $W$ does not
satisfy \cA.
\QED

\subsection{Some properties of preferential entailment}
\label{subsec:some}
The following result, Theorem~\ref{the:new}, is new. 
It is important for several reasons.
It uses the semantic representation of Theorem~\ref{compth:pref}
and a direct proof using only proof-theoretic arguments seems difficult.
It will be used in Section~\ref{subsec:rankpref}. 
Its Corollary~\ref{co:reso} 
should provide a starting point for the application to preferential entailment
of methods based on or related to resolution. 
First a definition.

\begin{definition}
\label{def:cons}
If a formula $\alpha$ is such that \mbox{$\alpha$ \notNI {\bf false}}, we shall
say that $\alpha$ is consistent (for the consequence relation \NI).
A formula is consistent for a model iff it is consistent for the 
consequence relation defined by the model, or equivalently iff there is
a state in the model that satisfies \ga.
\end{definition}
We shall now define a basic operation on preferential models.
Suppose $M$ is a preferential model \mbox{$\langle V , l , \prec \rangle$}.
For \mbox{$s , t \in V$} we shall write \mbox{$s \preceq t$} iff 
\mbox{$s \prec t$} or \mbox{$s = t$}.
Let $\alpha$ be a formula and $u \in V$ be a minimal element of 
$\widehat\alpha$.
Let $\prec_{\alpha}^{u}$ 
be the strict partial order obtained from $\prec$ by making $u$
a minimum of $\widehat\alpha$, i.e.,
\mbox{$s \prec_{\alpha}^{u} t$} iff \mbox{$s \prec t$} or 
\mbox{$s \preceq u$} and
there exists a state 
\mbox{$w \in \widehat\alpha$} such that \mbox{$w \preceq t$}.
The following lemma describes the properties of the construction
described above.

\begin{lemma}
\label{prelim}
The structure 
\mbox{$M_{\alpha}^{u} \eqdef \langle V , l , \prec_{\alpha}^{u} \rangle$}
is a preferential model. The consequence relation defined by $M_{\alpha}^{u}$ 
extends the consequence relation defined by $M$. In this model 
$u$ is a minimum of $\widehat\alpha$. Both models have the same set
of consistent formulas.
\end{lemma}
\proof
It is easy to see that $\prec_{\alpha}^{u}$ is irreflexive and transitive.
It is also easy to see that, under $\prec_{\alpha}^{u}$, 
$u$ is a minimum of $\widehat\alpha$.
We want to show now that, for any $\beta \in L$, 
the set
$\widehat\beta$ is smooth, under $\prec_{\alpha}^{u}$.
Let $s \in \widehat\beta$. Since $\widehat\beta$ is smooth 
under $\prec$, there is
a state $t$, minimal under $\prec$ in $\widehat\beta$ such that 
\mbox{$t \preceq s$}.
If $t$ is still minimal in $\widehat\beta$ under $\prec_{\alpha}^{u}$, 
then we are done.
If not, there is some state $v \in \widehat\beta$ such that
\mbox{$v \preceq u$} and \mbox{$v \prec_{\alpha}^{u} s$}. 
Since $\widehat\beta$ is smooth under $\prec$, there is a state $w$, minimal
in $\widehat\beta$ under $\prec$ such that \mbox{$w \prec v$}.
Since \mbox{$w \prec u$}, $w$ must be minimal in $\widehat\beta$ also under
$\prec_{\alpha}^{u}$. But \mbox{$w \prec_{\alpha}^{u} s$}.
We have shown that $\widehat\beta$ is smooth under $\prec_{\alpha}^{u}$.
To see that the consequence relation defined by $M_{\alpha}^{u}$ extends
the one defined by $M$, just notice that, since $\prec_{\alpha}^{u}$ extends
$\prec$, all minimal elements under the former are also minimal under the 
latter.
Lastly, since $M$ and $M_{\alpha}^{u}$ have exactly the
same set of worlds and the same labeling function, 
they define exactly the same set of consistent formulas.
\QED

\begin{theorem}
\label{the:new}
Let \bK\ be a knowledge base and \ab\ an assertion that is {\em not}
preferentially entailed by \bK. The formulas that are inconsistent
for the preferential closure of \/
\mbox{${\bf K} \cup \{ \alpha \NI \neg \beta \}$} are 
those that are inconsistent for the preferential closure of \bK.
\end{theorem}
\proof
Suppose that \ab\ is not preferentially entailed by \bK.
Then, let
\mbox{$W = \langle S , l, \prec \rangle$} 
be the preferential model the existence of which is guaranteed by
Theorem~\ref{compth:pref} and that defines \/${\bf K}^p$.
The model $W$ does not satisfy \ab. There is therefore a minimal element
$s \in S$ of $\widehat{\alpha}$ that does not satisfy $\beta$.
Consider now the model $W' \eqdef W_{\alpha}^{s}$. By Lemma~\ref{prelim}
this is a preferential model that satisfies all the assertions
satisfied by $W$, therefore it satisfies all the assertions of
\bK. Since $s$ is the only minimal element of $\widehat{\alpha}$,
it satisfies \mbox{$\bK \cup \{ \alpha \NIm \neg \beta \}$}.
Suppose \gc\ is inconsistent for
\mbox{$(\bK \cup \{ \alpha \NI \neg \beta \})^p$}. 
Then it must be inconsistent for $W'$. By Lemma~\ref{prelim}
it is inconsistent for $W$, therefore inconsistent for \/${\bf K}^p$.
\QED

\begin{corollary}
\label{co:reso}
Let \bK\ be a conditional knowledge base and \ab\ a conditional
assertion. The assertion \ab\ is preferentially entailed by \bK\
iff the assertion \mbox{\ga \NI {\bf false}} is preferentially
entailed by \mbox{$\bK \cup \{ \alpha \NIm \neg \beta \}$}. 
\end{corollary}
\proof
The {\em only if} part follows immediately from the soundness of
the {\bf And} rule.
The {\em if} part, follows immediately from Theorem~\ref{the:new}.
\QED

\subsection{The rank of a formula}
\label{subsec:ranking}
In this section, we introduce a powerful tool for studying preferential
entailment. 
Given a knowledge base, we shall attach an ordinal, its rank, to every formula.
We shall prove an important result concerning those ranks, and, in particular,
show that a knowledge base \bK\ and its preferential closure \bKp\ define
the same ranks.

\begin{definition}
\label{def:exc}
Let \bK\ be a conditional knowledge base (i.e., a set of conditional 
assertions) and \ga\ a formula. The formula \ga\ is said to be
{\em exceptional} for \bK\ iff \bK\ preferentially entails the assertion 
\mbox{{\bf true} \NI $\neg \alpha$}.
The conditional assertion \mbox{\cA $\eqdef$ \ga \NI \gb} is said to be
exceptional for \bK\ iff its antecedent \ga\ is exceptional for \bK.
\end{definition}
The set of all assertions of \bK\ that are exceptional for \bK\ will be
denoted by $E ( \bK )$. Notice that \mbox{$E(\bK) \subseteq \bK$}.
If all assertions of \bK\ are exceptional for \bK, i.e., if \bK\ is equal
to \mbox{$E(\bK)$}, we shall say that \bK\ is completely exceptional.
The empty knowledge base is completely exceptional.
Notice that, in the definition above, \bK\ may be replaced by 
its preferential closure \bKp.

Given a conditional knowledge base \bK\ (not necessarily finite),
we shall now define by ordinal induction an infinite non-increasing sequence
of subsets of \bK. 
Let $C_0$ be equal to \bK. For any successor ordinal $\tau + 1$,
$C_{\tau + 1}$ will be $E ( C_{\tau} )$ and for any limit ordinal
$\tau$, $C_{\tau}$ is the intersection of all $C_\rho$ for $\rho < \tau$.
It is clear that, after some point on, all $C$'s are equal and
completely exceptional (they may be empty, but need not be so).
We shall say that a formula \ga\ has rank $\tau$ (for \bK) 
iff $\tau$ is the least ordinal for which \ga\ is not exceptional for $C_\tau$.
A formula that is exceptional 
for all $C_\tau$'s is said to have no rank.
Notice that such a formula is exceptional for a totally exceptional
knowledge base.
The following is a fundamental lemma on preferential entailment.
It says that, as far as preferential entailment
is concerned, non-exceptional assertions cannot help deriving 
exceptional assertions. The notion of rank defined above proves to be
a powerful tool for studying preferential entailment.

\begin{lemma}
\label{le:funda}
Let $\tau$ be an ordinal.
Let \bK\ be a conditional knowledge base and \cA\ a conditional assertion
whose antecedent has rank larger or equal to $\tau$ (or has no rank).
Then \cA\ is preferentially entailed by $C_{0}$
iff it is preferentially entailed by $C_{\tau}$.
\end{lemma}
\proof
The {\em if} part follows from the fact that $C_{\tau}$ is a subset of
$C_{0}$.
The {\em only if} part is proved by induction on the length of the
proof of \cA\ from $C_{0}$.
If the proof has length one, i.e., \cA\ is obtained by {\bf Reflexivity}
or is an assertion of $C_{0}$, then the result is obvious.
If the last step of the proof is obtained by 
{\bf Right Weakening} or {\bf And}, the result follows from a trivial use
of the induction hypothesis.
If the last step of the proof is obtained by {\bf Left Logical Equivalence},
the result follows from the induction hypothesis and the fact that,
if \ga\ and $\alpha'$ are logically equivalent then \ga\ 
and $\alpha'$ have the same rank.
If the last step is a use of {\bf Or}, and \cA\ is of the form
\mbox{$\alpha \vee \beta$ \NI \gc} then just remark that
the rank of the disjunction \mbox{$\alpha \vee \beta$}
is the smaller of the ranks of \ga\ and \gb.
Both \ga\ and \gb\ have therefore a rank larger or equal to $\tau$
and one concludes by the induction hypothesis.
If the last step is a use of {\bf Cautious Monotonicity}, 
and \cA\ is of the form
\mbox{$\alpha \wedge \beta$ \NI \gc}, where \ab\ and \mbox{\ga \NI \gc}
are preferentially entailed (with short proofs) by $C_{0}$,
let $\sigma$ be the rank of \ga.
By the induction hypothesis $C_{\sigma}$ preferentially entails \ab.
Since \ga\ is not exceptional for $C_\sigma$, we conclude
that \mbox{$\alpha \wedge \beta$} is not exceptional for $C_{\sigma}$,
and therefore has rank $\sigma$.
But \mbox{$\alpha \wedge \beta$} has rank larger or equal to $\tau$.
Therefore \mbox{$\tau \leq \sigma$}.
The formula \ga\ has rank larger or equal to $\tau$ and we may apply
the induction hypothesis to conclude that both \ab\ and \mbox{\ga \NI \gc}
are preferentially entailed by $C_{\rho}$.
\QED

\begin{lemma}
\label{le:dontcare}
Let \bK\ and $\bK'$ be knowledge bases such that
\mbox{$\bK \subseteq \bK' \subseteq \bK^{p}$}.
For any formula, the rank it is given by $\bK'$ is equal to the rank it
is given by \bK.
\end{lemma}
\proof
Using Lemma~\ref{le:funda}, one shows by ordinal induction that 
\mbox{$C_{\tau} \subseteq C'_{\tau} \subseteq (C_{\tau})^p$}.
\QED
The following definition will be useful in Section~\ref{subsec:proof}.
\begin{definition}
\label{def:acc}
A knowledge base \bK\ is said to be {\em admissible} iff 
all formulas that have no rank for \bK\ are inconsistent for \bK.
\end{definition}
We shall immediately show that many knowledge bases are admissible.
\begin{lemma}
\label{co:acc}
If the preferential closure of \bK\ is defined by some 
well-founded preferential model, then
\bK\ is admissible.
In particular, any finite knowledge base is admissible.
\end{lemma}

\proof
We have noticed, in Lemma~\ref{le:dontcare} that ranks are stable 
under the the replacement of a knowledge base by its preferential closure.
Let $P$ be the preferential closure of \bK.
Suppose $P$ is defined by some well-founded preferential model $W$.
Suppose \ga\ has no rank. 
We must show that no state of $W$ satisfies \ga.
As noticed above, there is an ordinal $\tau$
such that $C_{\tau}$ is completely exceptional and \ga\ is exceptional
for $C_{\tau}$.
We shall show that no state of $W$ satisfies a formula that is exceptional 
for $C_{\tau}$.
Indeed, if there were such a state, there would be such a minimal state,
$s$, since $W$ is well-founded.
But $W$ is a model of $C_{\tau}$ and no state below $s$ satisfy any antecedent
of $C_{\tau}$, since $C_{\tau}$ is totally exceptional. 
Therefore the preferential model consisting of $s$ alone is a model of 
$C_{\tau}$.
But, in a model of $C_{\tau}$,
no minimal state satisfy a formula that is exceptional for $C_{\tau}$.
A contradiction.
It follows now from Lemma~\ref{le:co2} that any finite knowledge base is
admissible.
\QED

\subsection{Computing preferential entailment}
\label{subsec:comppref}
This section is devoted to the study of the computational complexity 
of preferential entailment. It is not needed in the sequel.
We shall explain in Section~\ref{subsec:discpref} why preferential 
entailment is not the 
{\em right} notion of entailment to answer the question of the title,
nevertheless preferential entailment is a central concept and it is
therefore worthwhile studying its computational complexity.
The results here are quite encouraging: the problem is in co-NP,
i.e., in the same polynomial class as the problem of deciding whether
a propositional formula is valid.

\begin{lemma}
\label{le:aux1}
Let \bK\ be a finite conditional knowledge base and \ab\ a conditional
assertion that is not preferentially entailed by \bK. 
There is a finite
totally (i.e., linearly) ordered preferential model of \bK\, 
no state of which satisfies \ga\ except the top state. This top state
satisfies \ga\ and does not satisfy \gb.
\end{lemma}
\proof
Let ${\cal L'} \subseteq {\cal L}$ be a logically finite language, 
large enough to contain
\ga, \gb\ and all the assertions of \bK.
Let us now consider ${\cal L'}$ to be our language of reference.
Clearly, \ab\ is not preferentially entailed by \bK, since a proof over
the smaller language is a proof over the larger language.
By Theorem~\ref{compth:pref}, there is a finite preferential model 
$W$ (over ${\cal L'}$) of \bK\ that does not satisfy \ab. 
In $W$, there is therefore a state $s$,
minimal in $\widehat{\alpha}$, 
that satisfies \ga\ but does not satisfy \gb.
Consider the submodel $W'$ obtained by deleting all states of $W$
that are not below or equal to $s$. 
It is clearly a finite preferential model of \bK, with a top state that
satisfies \ga\ but not \gb.
Let $V$ be obtained by imposing on the states of $W'$ any total ordering that
respects the partial ordering of $W'$. Since there are only finitely
many states in $V$, the smoothness condition is verified and $V$ is
a preferential model (on ${\cal L'}$). 
It is a model of \bK\ but not of \cA.
Now we may extend the labeling function of $V$ to the propositional
variables of ${\cal L}$ that are not in ${\cal L'}$ any way we want, to get
the model requested. Notice that the model obtained satisfies the
smoothness condition because it is finite.
\QED

\begin{theorem}
\label{comp}
There is a non-deterministic algorithm that,
given a finite set \/{\bf K} of conditional assertions and a conditional
assertion ${\cal A}$, checks that ${\cal A}$ is not preferentially
entailed by \/{\bf K}.
The running time of this algorithm is polynomial in the size of 
\/{\bf K} (sum of the sizes of its elements) and ${\cal A}$.
\end{theorem}
\proof
Let {\bf K} be 
\mbox{${\{ \gamma_i \NIm \delta_i \}}_{i = 1}^{N}$}.
Let 
\mbox{$I \subseteq \{1 , \ldots , N \}$} be a set of indices.
We shall define:
\mbox{$\varphi_{I} \eqdef \bigvee_{i \in I} \gamma_i$}
and
\mbox{$\psi_{I} \eqdef \bigwedge_{i \in I} \left ( \gamma_i 
\rightarrow \delta_i \right )$}.
A sequence is a sequence of pairs 
\mbox{$\left ( I_i , f_i \right )$}
for \mbox{$i = 0 , \ldots , n$}, where $I_i \subseteq I$
and $f_i$ is a world.
Let $\alpha$ and $\beta$ be in ${\cal L}$.
\begin{definition}
A sequence 
\mbox{$\left ( I_i , f_i \right )$}, \mbox{$i = 0 , \ldots , n$},
is a witness for \mbox{$\alpha$ \NI $\beta$} 
(we mean a witness that \mbox{$\alpha$ \NI $\beta$} 
is not preferentially 
entailed by \/{\bf K} ) iff
\begin{enumerate}
\item 
\label{it1}
\mbox{$f_{k} \models \psi_{I_{k}}$}, \ 
\mbox{$\forall k = 0 , \ldots , n$} 
\item
\label{itp}
\mbox{$f_{k} \models \varphi_{I_{k}}$}, \ 
\mbox{$\forall k = 0 , \ldots , n - 1$}
\item 
\label{it2}
\mbox{$I_{k + 1} = I_{k} \bigcap \{ j \mid f_{k} \not \models
\gamma_{j} \}$}, \ 
\mbox{$\forall k = 0 , \ldots , n - 1$}
\item 
\label{it3}
\mbox{$I_0 = \{1, \ldots , N \}$}
\item 
\label{it4}
\mbox{$f_{k} \not \models \alpha$}, \ 
\mbox{$\forall k = 0 , \ldots , n - 1$}
\item 
\label{it5}
\mbox{$f_{n} \models \alpha \wedge \neg \beta$}.
\end{enumerate}
\end{definition}
We must check that: witnesses are short and a conditional assertion
has a witness iff it is not preferentially entailed by {\bf K}.
For the first point, just remark that, for
\mbox{$k = 0 , \ldots , n - 1$} the inclusion
\mbox{$I_{k} \supset I_{k + 1}$} is strict because of 
items~\ref{it2} and \ref{itp}. The length of the sequence is therefore 
bounded by the number of assertions in {\bf K}. 
But, each pair has a short description. 
For the second point, suppose first there is a witness for
\mbox{$\alpha$ \NI $\beta$}. 
Then the ranked model $W$ consisting
of worlds $f_0 , \ldots , f_n$  where 
\mbox{$f_k \prec f_{k + 1}$} for
\mbox{$k = 0 , \ldots , n - 1$}
satisfies {\bf K} but not 
\mbox{$\alpha$ \NI $\beta$}.
That it does not satisfy 
\mbox{$\alpha$ \NI $\beta$}
is clear from items~\ref{it4} and \ref{it5}.
Let us check that $W$ satisfies
\mbox{$\gamma_i$ \NI $\delta_i$}.
If none of the $f_k$'s, \mbox{$k = 0 , \ldots , n$}
satisfies $\gamma_i$ then $W$ satisfies 
\mbox{$\gamma_i$ \NI $\eta$} for any $\eta$ in ${\cal L}$.
Suppose therefore that $j$ is the smallest $k$ for
which 
\mbox{$f_j \models \gamma_i$}. 
We must show that \mbox{$f_j \models \delta_i$}.
But, by items~\ref{it3} and \ref{it2}
\mbox{$i \in I_j$} and by item~\ref{it1},
\mbox{$f_i \models \delta_i$}.

Suppose now that
\mbox{$\alpha$ \NI $\beta$} is not preferentially entailed by some
given finite {\bf K}.
By Lemma~\ref{le:aux1},
there is a finite linearly ordered model $W$ of {\bf K}, 
no state of which satisfies $\alpha$, except the top state that is 
labeled by a world $m$ that
satisfies \mbox{$\alpha \wedge \neg \beta$}.
Let 
\mbox{$I_0 \eqdef \{1 , \ldots , N \}$}.
It is easy to see that ({\em remark} 1):
if $V$ is any preferential model of {\bf K}, for any set 
\mbox{$I \subseteq I_0$}, $V$ satisfies
\mbox{$\varphi_{I} \NI \psi_I$}.
Let us now consider first the set
\mbox{$\widehat{\alpha \vee \varphi_{I_{0}}}$}.
It cannot be empty, therefore it has a unique minimal state.
Let $f_0$ be the label of this state.
We must consider two cases.
First suppose that 
\mbox{$f_0 \models \alpha$}.
Then $f_0$ is minimal in $\widehat\alpha$ and therefore must be $m$.
In such a case
\mbox{$\left ( I_0 , m \right )$} is a witness.
The only thing to check is that item~\ref{it1} is satisfied.
Indeed either 
\mbox{$m \models \varphi_{I_{0}}$} and we conclude by remark~1
or 
\mbox{$\widehat {\varphi_{I_{0}}} = \emptyset$} and $m$ satisfies none of
the $\gamma_i$'s.
Let us deal now with the case 
\mbox{$f_0 \not \models \alpha$}.
We shall build a sequence beginning by
\mbox{$\left ( I_0 , m \right )$}.
Since $m$ does not satisfy $\alpha$, it must satisfy $\varphi_{I_{0}}$,
which takes care of item~\ref{itp}. Remark~1 takes care
of item~\ref{it1}. Let us now define
\mbox{$I_{1} = I_{0} \bigcap \{ j \mid f_{0} \not \models
\gamma_{j} \}$}. $I_1$ is strictly smaller than $I_0$.
We may now consider the set 
\mbox{$\widehat{\alpha \vee \varphi_{I_{1}}}$}.
It is not empty and therefore has a unique minimal element and
we may, in this way, go on and build a proof for
\mbox{$\alpha$ \NI $\beta$}.
\QED
Since it is clear that preferential non-entailment is at least as hard as
satisfiability (consider assertions with antecedent {\bf true}),
we conclude that it is an NP-complete problem, i.e., that preferential
entailment is co-NP-complete.
A remark of J.~Dix that will be explained at the end of
Section~\ref{subsec:comprat} shows that preferential entailment is reducible
to the computation of rational closure and that this reduction, when applied
to Horn formulas, requires only the consideration of Horn formulas.
It follows that, if we restrict ourselves to {\em Horn} assertions,
computing preferential entailment has only polynomial complexity.

\section{Rationality}
\label{sec:rat}
\subsection{Introduction}
\label{subsec:introrat}
In this section we explain why not all preferential relations represent
reasonable nonmonotonic inference procedures.
We present some additional principles of nonmonotonic reasoning and discuss 
them.
Those principles are structurally different from the rules of preferential 
reasoning, since they are not of the type:
deduce some assertion from some other assertions.
Sections~\ref{subsec:negrat} and~\ref{subsec:disrat} present
weak principles. Some results are proven concerning those principles.
Deeper results on those principles, found after a first version of this paper
had been circulated, appear in~\cite{FLMo:91}.
Our central principle is presented in Section~\ref{subsec:ratmon}.
Those principles were first described in~\cite{KLMAI:89} but
the technical results presented here are new.
In~\ref{subsec:discpref}, the value of preferential entailment as an answer
to the question of the title is discussed.
Our conclusion is that it is not a satisfactory answer, 
since it does not provide us with a 
rational relation.
Then, in Section~\ref{subsec:rank},
a restricted family of preferential models, the family of ranked models, 
is presented and a representation theorem is proved.
The result is central to this paper but the proof of the representation
theorem may be skipped on a first reading.  
The representation theorem appeared in~\cite{LMTR:88}.
The family of ranked models is closely related to, but different from,
a family studied in~\cite{Del:87} and Section~\ref{subsec:del} explains
the differences.

\subsection{Negation Rationality}
\label{subsec:negrat}
In~\cite[Section 5.4]{KLMAI:89}, it was argued that not all preferential
consequence relations represented reasonable inference operations.
Three {\em rationality} properties were discussed there, and it was
argued that all three were desirable. 
Those properties do not lend themselves to be presented as {\em Horn}
rules (deduce the presence of an assertion in a relation from 
the presence of other assertions) but have the form: deduce the absence
of an assertion from the absence of other assertions. All of them
are implied by {\bf Monotonicity}.
The reader may find the discussion of~\cite{KLMAI:89} useful. 
Here technical results will be described.
The first property considered is the following.
\begin{equation}
\label{neg:rat}
{{\alpha \wedge \gamma \notNIm \beta \ , 
\ \alpha \wedge \neg \gamma \notNIm \beta} 
\over
{\alpha \notNIm \beta}} \hspace {1.6cm} 
{\rm ({\bf Negation \ Rationality}) }
\end{equation}
\begin{lemma}
\label{le:neg:rat}
There is a preferential relation that does not satisfy 
{\bf Negation Rationality}.
\end{lemma}
\proof
Take a preferential model containing four states: 
\mbox{$s_i , i = 0 , \ldots 3$}, with $s_0 \prec s_1$
and $s_2 \prec s_3$. Let the even states be the only states satisfying $q$
and $s_0$ and $s_3$ be the only states satisfying $p$. One easily
verifies that the consequence relation defined by this model is
such that 
\mbox{{\bf true} \NI $q$}, but 
\mbox{$p$ \notNI $q$} and
\mbox{$\neg p$ \notNI $q$}.
\QED
No semantic characterization of relations satisfying {\bf Negation Rationality}
is known.
It has been shown in~\cite{KLMAI:89} that the consequence relation
defined by {\em Circumscription} does not always satisfy 
{\bf Negation Rationality}.

\subsection{Disjunctive Rationality}
\label{subsec:disrat}
The next property is the following.
\begin{equation}
\label{disj:rat}
{{\alpha \notNIm \gamma \ \ , \ \ \beta \notNIm \gamma}
\over {\alpha \vee \beta \notNIm \gamma}} \hspace {3cm}
{\rm ({\bf Disjunctive \ Rationality}) }
\end{equation}
We may prove the following.
\begin{lemma}
\label{le:disneg}
Any preferential relation that satisfies {\bf Disjunctive Rationality}
satisfies {\bf Negation Rationality}.
\end{lemma}
\proof
Suppose \mbox{$\alpha \wedge \gamma$ \notNI \gb}
and \mbox{$\alpha \wedge \neg \gamma$ \notNI \gb}.
By {\bf Disjunctive Rationality}, we conclude that
\mbox{$\alpha \wedge \gamma \vee \alpha \wedge \neg \gamma$ \notNI \gb}.
We conclude by {\bf Left Logical Equivalence}.
\QED
\begin{lemma}
\label{le:disneg2}
There is a preferential relation that satisfies {\bf Negation Rationality}
but does not satisfy {\bf Disjunctive Rationality}.
\end{lemma}
\proof
Let us consider the following preferential model $W$.
The model $W$ has four states: $a_0 , a_1 , b_0 , b_1$. 
The ordering is: \mbox{$a_0 \prec a_1$} and 
\mbox{$b_0 \prec b_1$}. 
The language has three propositional
variables: $p$, $q$ and $r$. 
The two states $a_1$ and $b_1$ (the top states) are labeled
with the same world that satisfies only $p$ and $q$. 
State $a_0$ is labeled with the world that satisfies only $p$ and $r$
and the state $b_0$ with the world that satisfies only $q$ and $r$.
The preferential relation defined by $W$ does not satisfy {\bf Disjunctive 
Rationality} but satisfies {\bf Weak Rationality}.
For the first claim, notice that: 
\mbox{$p \vee q$ \NIW $r$} but 
\mbox{$p$ \notNIW $r$} and
\mbox{$q$ \notNIW $r$}.
For the second claim, 
suppose \mbox{\ga \NIW \gc}, but 
\mbox{$\alpha \wedge \beta$ \notNIW $\gc$}.
Then it must be the case that there is a minimal state of $\widehat{\alpha}$ 
that does not satisfy \gb\ and, above it, 
a state that is minimal in $\widehat{\alpha \wedge \beta}$. 
This last state must be labeled by a world that is the label of 
no minimal state of $\widehat{\alpha}$. 
Therefore, $\widehat{\alpha}$ must contain all four states of $W$,
and $\widehat{\alpha \wedge \beta}$ must contain either 
the two top states alone or the two top states and one of the bottom states. 
In each case it is easy to see that \mbox{$\alpha \wedge \neg \beta$ \NIW \gc}
since the minimal states of $\widehat{\alpha \wedge \neg \beta}$ 
are all also minimal in $\widehat{\alpha}$.
\QED
No semantic characterization of relations satisfying 
{\bf Disjunctive Rationality} was known at the time this paper was elaborated.
M.~Freund~\cite{Freund:91} has now provided a very elegant
such characterization
together with an alternative proof of our Theorem~\ref{comthe:rat}; the
canonical model he builds is essentially the same as ours.

\subsection{Rational Monotonicity}
\label{subsec:ratmon}
The last property is the following.
\begin{equation}
\label{Rat:mon}
{{\alpha \wedge \beta \notNIm \gamma \ \ , \ \ \alpha \notNIm \neg\beta} 
\over
{\alpha \notNIm \gamma}} \hspace {2cm}
{\rm ({\bf Rational \ Monotonicity}) }
\end{equation}
This rule is similar to the thesis {\bf CV} of conditional logic 
(see~\cite{Nute:84}).
The reader is referred to~\cite[Section 5.4]{KLMAI:89} for a discussion
of our claim that reasonable consequence relations should 
satisfy {\bf Rational Monotonicity}.
Some researchers in Conditional Logic (J.~Pollock in particular)
have objected to {\bf CV} as a valid thesis for (mainly subjunctive) 
conditionals. 
Echoes of this debate may be found in~\cite[end of Section 4.4]{Gin:86}.
The objections to {\bf CV} that hold in the conditional logic framework
do not hold for us, though their consideration is recommended to the reader.
The most attractive feature of {\bf Rational Monotonicity} is probably that
it says that an agent should not have to retract any previous defeasible
conclusion when learning about a new fact the negation of which was not 
previously derivable. In~\cite{Satoh:89}, K.~Satoh aptly decided
to call nonmonotonic reasoning that validates {\bf Rational Monotonicity}
{\em lazy}.
The rule of {\bf Rational Monotonicity} should be distinguished from 
the following rule, which is satisfied by any preferential relation.
\begin{equation}
\label{eq:prat}
{{\alpha \wedge \beta \NIm \neg \gamma \ \ , \ \ \alpha \notNIm \neg\beta} 
\over
{\alpha \notNIm \gamma}}
\end{equation}
\begin{definition}
\label{def:ratrel}
A rational consequence relation is a preferential relation that
satisfies {\bf Rational Monotonicity}.
\end{definition}
Two different representation theorems will be proved about
rational relations, in Sections~\ref{subsec:rank} 
and in Appendix~\ref{appen:nonstandard}.
The last one seems to provide evidence that reasonable inference 
procedures validate {\bf Rational Monotonicity} and that all rational 
relations represent reasonable inference procedures.
\begin{lemma}
\label{le:ratmondisj}
A rational relation satisfies {\bf Disjunctive Rationality}.
\end{lemma}
\proof
Suppose \mbox{\ga \notNI \gc} and \mbox{\gb \notNI \gc}.
By {\bf Left Logical Equivalence} we have
\mbox{$(\alpha \vee \beta ) \wedge \alpha$ \notNI \gc}.
If we have \mbox{$\alpha \vee \beta$ \notNI $\neg \alpha$},
then we could conclude by {\bf Rational Monotonicity} that
\mbox{$\alpha \vee \beta$ \notNI \gc}.
Suppose then that  \mbox{$\alpha \vee \beta$ \NI $\neg \alpha$}.
If we had \mbox{$\alpha \vee \beta$ \NI \gc}, we would
conclude by preferential reasoning that \mbox{\gb \notNI \gc}.
\QED
\begin{lemma}[David Makinson]
\label{DR<RM}
There is a preferential relation satisfying {\bf Disjunctive Rationality}
that is not rational.
\end{lemma}
\proof
We shall build a preferential model that defines a consequence relation 
satisfying {\bf Disjunctive Rationality} but not {\bf Rational Monotonicity}.
Let ${\cal L}$ be the propositional calculus on the three variables: $p_{0}$,
$p_{1}$, $p_{2}$. 
Let ${\cal U}$ contain all propositional worlds
on those variables.
Let $S$ contain three elements: $s_i$ for $i = 0 , 1 , 2$ and
$l(s_i)$ satisfy only $p_i$. The partial order $\prec$ is such that
$s_1 \prec s_2$ and no other pair satisfies the relation.
This defines a preferential model $W$.
First we shall show that the consequence relation defined by $W$ does 
not satisfy {\bf Rational Monotonicity}. 
Indeed, we have both \mbox{$p_0 \vee p_1 \vee p_2$ \NIW $\neg p_2$} and
\mbox{$p_0 \vee p_1 \vee p_2$ \notNIW $p_1$}.
Nevertheless, we also have
\mbox{$\neg p_1 \wedge ( p_0 \vee p_1 \vee p_2 )$ \notNIW $\neg p_2$}.
Let us show now that any preferential model that does not satisfy
{\bf Disjunctive Rationality} must have at least $4$ states.
Suppose 
\mbox{$\alpha \vee \beta$ \NI $\gamma$}, but
\mbox{$\alpha$ \notNI $\gamma$} and
\mbox{$\beta$ \notNI $\gamma$}.
The last two assumptions imply the existence of states $a$ and $b$, 
minimal in $\widehat\alpha$ and $\widehat\beta$ respectively and that do not
satisfy $\gamma$, and therefore are not minimal in
\mbox{$\widehat{\alpha \vee \beta}$}. 
Those states are different, since any state minimal in 
both $\widehat\alpha$ and $\widehat\beta$ would be minimal in 
\mbox{$\widehat{\alpha \vee \beta}$}.
By the smoothness condition there must be a state $a'$ minimal in 
$\widehat{\alpha \vee \beta}$ and such that
$a' \prec a$. Clearly $a'$ satisfies $\gamma$ and does not satisfy 
$\alpha$ (since $a$ is minimal in
$\widehat\alpha$) but satisfies $\beta$.
Similarly there must be a state $b'$ minimal in
$\widehat{\alpha \vee \beta}$ and such that
$b' \prec b$ and $b'$ satisfies $\gamma$ and does not satisfy $\beta$,
but satisfies $\alpha$.
It is left to show that all four states are different.
We have already noticed that $a \not = b$. The states $a'$ and $b'$ satisfy
$\gamma$ and are therefore different from $a$ and $b$. 
But $b'$ satisfies
$\alpha$ and $a'$ does not and therefore $a' \not = b'$.
\QED

\subsection{Discussion of Preferential Entailment}
\label{subsec:discpref}
We may now assess preferential entailment as a possible answer
to the question of the title.
Corollary~\ref{co1} explains why the notion of preferential
entailment cannot be the one we are looking for: the relation
${\bf K}^p$ can be any preferential relation and is not in general 
rational. For typical {\bf K}'s, ${\bf K}^p$ fails to satisfy a large
number of instances of {\bf Rational Monotonicity} and is therefore
highly unsuitable.
One particularly annoying instance of this is the following.
Suppose a conditional knowledge base {\bf K} contains one single
assertion 
\mbox{$p$ \NI $q$} 
where $p$ and $q$ are different propositional
variables. Let $r$ be a propositional variable, different from
$p$ and $q$. We intuitively expect the assertion
\mbox{$p \wedge r$ \NI $q$} 
to follow from {\bf K}. The rationale for that
has been discussed extensively in the literature and boils down to this:
since we have no information whatsoever about the influence of $r$ on
objects satisfying $p$ it is sensible to assume that it has no influence and
that there are normal
$p$-objects that satisfy $r$. The normal $p \wedge r$-objects are therefore
normal $p$-objects and have all the properties enjoyed by normal
$p$-objects.
Nevertheless it is easy to check that 
\mbox{$p \wedge r$ \NI $q$} 
is not in ${\bf K}^p$. 
The problem lies, at least in part, with the fact that ${\bf K}^p$ is not 
rational, since any
rational relation containing
\mbox{$p$ \NI $q$} ,
must contain 
\mbox{$p \wedge r$ \NI $q$} 
unless it contains
\mbox{$p$ \NI $\neg r$}.

In conclusion, it seems that the set of conditional assertions 
entailed by {\bf K} should be larger and more {\em monotonic} 
than the set \/${\bf K}^p$. It should also be rational.
This question will be brought up again in Section~\ref{sec:ratclos} 
and a solution
will be proposed.

\subsection{Ranked models and a representation theorem for rational relations}
\label{subsec:rank}
In this section a family of preferential models will be defined
and it will be shown that the relations defined by models of this
family are exactly the rational relations.
\begin{lemma}
\label{le:modular}
If $\prec$ is a partial order on a set $V$, the following conditions
are equivalent.
\begin{enumerate}
\item for any \mbox{$x , y , z \in V$} if \mbox{$x \not \prec y$},
\mbox{$y \not \prec x$} and \mbox{$z \prec x$}, then \mbox{$z \prec y$}
\item  for any \mbox{$x , y , z \in V$} if \mbox{$x \prec y$}, then, either
\mbox{$z \prec y$} or \mbox{$x \prec z$}
\item for any \mbox{$x , y , z \in V$} if \mbox{$x \not \prec y$} and 
\mbox{$y \not \prec z$}, then \mbox{$x \not \prec z$} 
\item \label{it30} there is a totally ordered set $\Omega$ 
(the strict order on $\Omega$ will be denoted by $<$) 
and a function \mbox{$r : V \mapsto \Omega$} (the ranking function) 
such that 
\mbox{$s \prec t$} iff \mbox{$r(s) < r(t)$}.
\end{enumerate}
\end{lemma}
The proof is simple and will not be given.
A partial order satisfying any of the conditions of Lemma~\ref{le:modular}
will be called {\em modular}
(this terminology is proposed in~\cite{Gin:86} as an extension of the
notion of modular lattice of~\cite{Gra:71}).
\begin{definition}
\label{rankmod}
A ranked model $W$ is a preferential model 
\mbox{$\langle V , l , \prec \rangle$} for which the strict partial order
$\prec$ is modular.
\end{definition}
Those models are called ranked since
the effect of function $r$ of property~\ref{it30} of 
Lemma~\ref{le:modular} is to rank the states: a state of smaller rank 
being more normal than a state of higher rank.
We shall always suppose that a ranked model $W$ comes equipped
with a totally ordered set $\Omega$ and a ranking function $r$.
Notice that we still require $W$ to satisfy the smoothness condition.
It is easy to see that for any subset $T$ of $V$ and any $t \in T$,
$t$ is minimal in $T$ iff $r(t)$ is the minimum of the set $r(T)$.
It follows that all minimal elements of $T$ have the same image by $r$.
The smoothness condition is then equivalent to
the following:
for any formula \mbox{$\alpha \in L$}, if $\widehat\alpha$ is not empty,
the set $r(\widehat\alpha)$ has a minimum. 
The smoothness condition is always verified if $\Omega$ is a well-ordered set.
The reader may check that the preferential model $W$ defined in the 
proof of Lemma~\ref{le:well} is ranked (it is even totally ordered).
It follows that there are rational relations that are defined by no
well-founded ranked model.
The following is a soundness result.
\begin{lemma}
\label{le:soundrat}
If $W$ is a ranked model, the consequence relation \NIW it defines
is rational.
\end{lemma}
\proof
It is enough to show that \NIW satisfies {\bf Rational Monotonicity}.
For this, the smoothness condition is not needed; it is needed,
though, for the soundness of {\bf Cautious Monotonicity}.
Suppose $W$ is a ranked model.
We shall use the notations of 
Definition~\ref{rankmod}. Suppose also that 
\mbox{$\alpha$ \NIW $\gamma$} 
and 
\mbox{$\alpha$ \notNIW $\neg\beta$}.
From this last assumption we conclude that there is 
a minimal element of $\widehat\alpha$ that satisfies $\beta$.
Let \mbox{$t \in V$} be such a state.
Let \mbox{$s \in V$} be a minimal element of 
\mbox{$\widehat{\alpha \wedge \beta}$}. 
Since \mbox{$t \in \widehat{\alpha \wedge \beta}$},
\mbox{$t \not \prec s$} and \mbox{$r(s) \leq r(t)$}.
But this implies that $s$ is minimal in $\widehat\alpha$: any state $u$
such that $u \prec s$ satisfies $r(u) < r(s)$ and therefore $r(u) < r(t)$
and $u \prec t$.
Since \mbox{$\alpha$ \NIW $\gamma$},
\mbox{$s \EM \gamma$}.
\QED
We shall show now that the converse of Lemma~\ref{le:soundrat} holds.
We shall first mention four derived rules of preferential logic.
In fact the first three of these rules are even valid in cumulative logic
(see~\cite[Section 3]{KLMAI:89}).
Their proof (either proof-theoretic or model-theoretic) is straightforward 
and is omitted.
\begin{lemma}
The following rules are derived rules of preferential logic:
\begin{equation}
\label{eq-1}
{{\alpha \NIm {\bf false}} \over
{\alpha \wedge \beta \NIm {\bf false}}}
\end{equation}
\begin{equation}
\label{eqzero}
{{\alpha \vee \beta \NIm \neg\beta} \over
{\alpha \NIm \neg\beta}}
\end{equation}
\begin{equation}
\label{eqtwo}
{{\alpha \vee \beta \vee \gamma \NIm \neg\alpha \wedge \neg\beta}
\over
{\beta \vee \gamma \NIm \neg\beta}}
\end{equation}
\begin{equation}
\label{eqone}
{{\alpha \vee \beta \NIm \neg\alpha} \over
{\alpha \vee \beta \vee \gamma \NIm \neg\alpha}}
\end{equation}
\end{lemma}

We shall now derive a property of rational relations.
\begin{lemma}
\label{rankder}
If \NI is a rational relation, then the following rule is valid:
\begin{equation}
\label{eqthree}
{{\alpha \vee \gamma \NIm \neg\alpha \ \  , \ \  \beta \vee \gamma \notNIm
\neg\beta} \over
{\alpha \vee \beta \NIm \neg\alpha}}
\end{equation}
\end{lemma}
\proof
From the first hypothesis, by Rule~\ref{eqone} one deduces 
\mbox{$\alpha \vee \beta \vee \gamma$ \NI $\neg\alpha$}.
From the second hypothesis, by Rule~\ref{eqtwo} one deduces
\mbox{$\alpha \vee \beta \vee \gamma$ \notNI $\neg ( \alpha \vee \beta )$}.
If one applies now {\bf Rational Monotonicity}, one gets the desired conclusion.
\QED
For the completeness result, we proceed in the style of L.~Henkin.
Completeness proofs in this style have been used in conditional logics
since~\cite{StalThom:70}. Since a number of technical lemmas are needed,
we have relegated them to Appendix~\ref{appen:rep} and state here 
the characterization theorem.
\begin{theorem}
\label{comthe:rat}
A binary relation \mbox{$\NI$} on ${\cal L}$ is a rational consequence 
relation 
iff it is the consequence relation defined by some ranked model.
If the language ${\cal L}$ is logically finite, then every rational consequence
relation is defined by some finite ranked model.
\end{theorem}
\proof
The {\em if} part is Lemma~\ref{le:soundrat}.
For the {\em only if} part, 
let \NI be a consequence relation satisfying the rules of {\bf R}.
The relation \NI defines a structure $W$  as described in 
Appendix~\ref{appen:rep}.
By Lemmas~\ref{smoothness} and \ref{total}, $W$ is a ranked model.
We claim that, for any \mbox{$\alpha , \beta \in L$}, 
\mbox{$\alpha$ \NI $\beta$} iff 
\mbox{$\alpha$ \NIW $\beta$}.
Suppose first \mbox{$\alpha$ \NI $\beta$}.
By Lemma~\ref{nottoomany}, if 
\mbox{$<m , \beta>$} 
is minimal in $\widehat\alpha$, then $m$ is normal 
for $\alpha$. We conclude that \mbox{$m \models \beta$} and
\mbox{$<m , \beta> \EM \beta$}. Therefore 
\mbox{$\alpha$ \NIW $\beta$}.
Suppose now that
\mbox{$\alpha$ \NIW $\beta$}.
Let $m$ be a normal world for $\alpha$. By Corollary~\ref{enough}, the pair 
\mbox{$<m , \alpha>$} is minimal in $\widehat\alpha$ and therefore
\mbox{$m \models \beta$}. All normal worlds for $\alpha$ therefore
satisfy $\beta$ and Lemma~8  of \cite{KLMAI:89} implies that
\mbox{$\alpha$ \NI $\beta$}.
For the last sentence of the theorem, notice that,
if the language ${\cal L}$ is logically finite, the model $W$ is finite.
\QED
As we remarked just prior to Lemma~\ref{le:soundrat}, the theorem would not 
hold had we required models to be well-founded.

\subsection{Comparison with Delgrande's work}
\label{subsec:del}
The system of proof-rules and the models presented above
may be compared with the results of 
J. Delgrande in~\cite{Del:87,Del:88}. The general thrust is very
similar but differences are worth noticing.
A first difference is in the language used.
Delgrande's language differs from this paper's in three respects: 
his work is specifically tailored to first-order predicate calculus, 
whereas this work deals with propositional calculus; 
he allows negation and disjunction of conditional assertions, 
which are not allowed in this paper;
he allows nesting of conditional operators in the language, 
though his completeness result is formulated only for unnested formulas.
Therefore Delgrande's central completeness result in \cite{Del:87}, 
only shows that any proposition in which there is no nesting 
of conditional operators (let us call those propositions {\em flat}) 
that is valid has a proof from the axioms and rules of his system. 
But this proof may use propositions that are not flat. 
The completeness results reported here show that valid assertions have
proofs that contain only flat assertions.

A second difference is that Delgrande's logical system is different from ours: 
Delgrande's logic {\bf N} does not contain {\bf Cautious Monotonicity}.
Our class of ranked models is more restricted than his class of models:
our models are required to obey the smoothness 
condition and Delgrande's are not. 
One may also notice that our logic enjoys
the finite model property, but Delgrande's does not. 
This difference between our two logical systems may sound insignificant 
when one
remarks that many instances of the rule of {\bf Cautious Monotonicity} may
be derived from {\bf Rational Monotonicity}, and are therefore valid in
Delgrande's system {\bf N}. 
What we mean is that if \mbox{$\alpha$ \NI $\beta$} and
\mbox{$\alpha$ \NI $\gamma$} then, if \mbox{$\alpha$ \notNI $\neg\beta$}
one may conclude \mbox{$\alpha \wedge \beta$ \NI $\gamma$} by {\bf Rational
Monotonicity} rather than by {\bf Cautious Monotonicity}. But if
\mbox{$\alpha$ \NI $\neg\beta$}, and therefore \mbox{$\alpha$ \NI {\bf false}}
one cannot conclude.
The Rule~\ref{eq-1} is sound in preferential logic but 
not in Delgrande's logic.
A proof will soon be given.
We want to remark here that Rule~(\ref{eq-1}) is very natural,
since the meaning of \mbox{$\alpha$ \NI {\bf false}} is that if $\alpha$ is
true than anything may be true. It therefore means that it is absolutely
unthinkable that $\alpha$ be true.
In such a case we would expect $\alpha \wedge \beta$ to be also absolutely
unthinkable.

Let us show now that Rule~(\ref{eq-1}) is not valid for Delgrande's structures.
Consider the following structure.
Let the set $V$ consists of one infinite 
descending chain: $\prec$ is a total ordering. 
Suppose now that the top element of $V$ is the only state that satisfies
the propositional variable $p$.
In this structure $\widehat{\bf true}$ is $V$ and has no minimal element,
therefore \mbox{${\bf true} \NI {\bf false}$}. But $\widehat p$ consists
only of the top element and has a minimal point and therefore 
\mbox{$p$ \notNI {\bf false}}.
We have shown that the Rule~\ref{eq-1} is not valid for Delgrande's
structures.
This example also shows that Delgrande's logic does not posses 
the finite model property. 

A third difference is that his definition of the set of conditional assertions
entailed by a conditional knowledge base is different from the one presented
here, at least at first sight.

\section{Ranked entailment}
\label{sec:entrank}
\subsection{Introduction}
\label{subsec:intrank}
After having defined the family of relations and the family of models
we are interested in, we proceed to study the notion of entailment 
provided by those models.
Our main result is presented in~\ref{subsec:rankpref}.
It is negative, in the sense that this entailment is equivalent to
preferential entailment.
A preliminary version of this result may be found in~\cite{LMTR:88}.
The collapsing of the two notions of entailment, as opposed to the two
different classes of relations represented, sheds a new light
on the results of~\cite{Adams:75}.
Section~\ref{subsec:adams} describes the probabilistic semantics
given to preferential entailment by E.~Adams in~\cite{Adams:75}
and shows how the result of Section~\ref{subsec:rankpref}
provides an alternative proof for Adams' results. 
The results of this section were contained in~\cite{LMTR:88}.

\subsection{Ranked entailment is preferential entailment}
\label{subsec:rankpref}
In the discussion of Section~\ref{subsec:discpref}, we expressed the wish
that the set of assertions entailed by a conditional knowledge base \bK\
be rational and larger than \/${\bf K}^p$. A natural candidate would be
the set of all assertions that are satisfied in all ranked models
that satisfy the assertions of \bK. This is an intersection of rational
relations.
This proposal fails in a spectacular way. Problems with this proposal
have been noted in~\cite[Section 4]{Del:88}.
It is also easy to see that the intersection of rational relations
may fail to be rational. Theorem~\ref{rat:ent} shows this failure to
be total.
\begin{lemma}
\label{le:auxrat}
Let \cE\ be any preferential relation. There exists a rational extension
of \cE\ for which a formula is inconsistent only if it is inconsistent
for \cE.
\end{lemma}
\proof
Let us choose some enumeration of triples of formulas
\ga, \gb\ and \gc\ in which every triple appears
an unbounded number of times.
Let $K_0$ be equal to \cE.
At every step $i$ we define $K_{i+1}$ in the following way.
Let \ga, \gb\ and \gc\ be the triple enumerated at step $i$.
Unless $K_i$ contains the pair \mbox{\ga \NI \gc} but contains
neither \mbox{$\alpha \wedge \beta$  \NI \gc} nor 
\mbox{\ga \NI $\neg \beta$}, we shall take $K_{i+1}$ to be equal
to $K_i$.
If $K_i$ satisfies the condition above, we shall take $K_{i+1}$
to be the preferential closure of \mbox{$K_{i} \cup \{ \ga \NI \gb \}$}.
Notice that, by {\bf Cautious Monotonicity}, 
\mbox{$\alpha \wedge \beta$  \NI \gc} will enter $K_{i+1}$.
It is clear that the $K_i$'s provide an increasing sequence of
preferential extensions of \cE. Let $K_\infty$ be the
union of all the $K_i$'s. Clearly $K_\infty$ is a preferential
extension of \cE.
By construction, and since we took care of removing all counter-examples
to the rule of {\bf Rational Monotonicity}, $K_\infty$ is a rational
consequence relation.
We claim that a formula \ga\ is inconsistent for $K_\infty$
(i.e., \mbox{\ga \NI {\bf false}} is in $K_\infty$) only if
it is inconsistent already for \cE.
Indeed, if \ga\ is inconsistent for $K_\infty$ it must be inconsistent
for some $K_i$, but
Theorem~\ref{the:new} shows that, by construction, all $K_i$'s have
the same inconsistent formulas.
\QED
\begin{theorem}
\label{rat:ent}
If the assertion \cA\ is satisfied by all {\em ranked} models that
satisfy all the assertions of {\bf K}, then it is satisfied by all
{\em preferential} such models.
\end{theorem}
\proof
Let \mbox{${\cal A} \eqdef \delta \NI \varepsilon$} be as in the hypotheses.
Let \cE\ be the rational extension of the preferential closure of 
\mbox{${\bf K} \cup \{ \delta \NI \neg \varepsilon \}$}, whose existence
is asserted by Lemma~\ref{le:auxrat}.
The assertion \mbox{$\delta \NI \varepsilon$} is in \cE\  
since it is in any rational relation that extends \bK, 
by Theorem~\ref{comthe:rat}.
Since $\delta \NI \neg \varepsilon$ is obviously in \cE, 
we conclude that $\delta$ is inconsistent for \cE\ 
and therefore inconsistent for the preferential closure of 
\mbox{${\bf K} \cup \{ \delta \NI \neg \varepsilon \}$}.
By Corollary~\ref{co:reso}, \cA\ is preferentially entailed by \bK.
\QED

\subsection{Comparison with Adams' probabilistic entailment}
\label{subsec:adams}
In this section we shall show that ranked models are closely related to
Adams' probabilistic entailment described in~\cite{Adams:75}.
Theorem~\ref{rat:ent}, then, provides an alternative proof of Adams' axiomatic
characterization of probabilistic entailment.
There are some technical differences between Adams' framework and ours
since Adams insists on allowing formulas as conditional assertions:
for him the formula $\alpha$ is a synonym for \mbox{{\bf true} \NI $\alpha$}.
We also insist on studying infinite knowledge bases whenever possible,
where Adams restricts himself to finite knowledge bases.

A probability assignment for the language ${\cal L}$ is a probability measure
on ${\cal L}$ yielded by some probability measure given on ${\cal U}$.
E.~Adams proposed the following definitions.
\begin{definition}[Adams]
\label{def:proper}
A probability assignment $p$ for the language ${\cal L}$ is said to be 
{\em proper}
for a conditional assertion \mbox{$\alpha$ \NI $\beta$} iff $p(\alpha) > 0$.
It is proper for a set of conditional assertions iff it is proper for
each element.
If $p$ is proper for \mbox{${\cal A} \eqdef \alpha$ \NI $\beta$}, we shall
use $p({\cal A})$ to denote the conditional probability $p(\beta \mid \alpha)$.
\end{definition}
\begin{definition}[Adams]
\label{def:pcons}
Let {\bf K} be a set of conditional assertions. We shall say that {\bf K}
is {\em probabilistically consistent} if and only if for any real number
\mbox{$\epsilon > 0$} there exists a probability assignment $p$ for ${\cal L}$ that
is proper for {\bf K} and such that, for all ${\cal A}$ in {\bf K}, one has
\mbox{$p({\cal A}) \geq 1 - \epsilon$}.
\end{definition}
\begin{definition}
\label{def:pentail}
Let {\bf K} be a set of conditional assertions and ${\cal A}$ a conditional
assertion. We shall say that {\bf K} {\em probabilistically entails}
${\cal A}$ iff for all $\epsilon > 0$ there exists $\delta >0$ such that
for all probability assignments $p$ for ${\cal L}$ which are proper for {\bf K}
and ${\cal A}$, if $p({\cal B}) \geq 1 - \delta$ for all ${\cal B}$
in {\bf K}, then $p({\cal A}) \geq 1 - \epsilon$.
\end{definition}
In \cite{Adams:75}, Adams studies extensively the relations between the two
notions of probabilistic consistency and probabilistic entailment, at least 
for finite sets of conditional assertions.
Here we shall only show the fundamental relation that exists between
Adams' notions and ours.
First, we shall make three easy remarks.
The first one concerns only probabilistic notions and was claimed by Adams 
for finite knowledge bases but is true in general.
\begin{lemma}
\label{pcon:pent}
A set {\bf K} of conditional assertions is probabilistically inconsistent
iff it probabilistically entails any conditional assertion.
\end{lemma}
Our second remark provides a first link between probabilistic notions
and the notions introduced in this paper. It is essentially the soundness
part of Adams' soundness and completeness result 
(see beginning of proof of 4.2 at page 62 of~\cite{Adams:75}).
This is the easy direction.
\begin{lemma}
\label{soundA}
Any conditional assertion preferentially entailed by {\bf K} is 
probabilistically entailed by {\bf K}.
\end{lemma}
Our third remark is the following.
\begin{lemma}
\label{133}
If the conditional assertion 
\mbox{$\alpha$ \NI $\beta$} is in {\bf K} and 
\mbox{$\alpha$ \NI {\bf false}} is preferentially entailed by {\bf K} 
then {\bf K}
is probabilistically inconsistent.
\end{lemma}
\proof
Under the assumptions of the lemma, Lemma~\ref{soundA} shows that 
\mbox{$\alpha$ \NI {\bf false}} is probabilistically entailed by {\bf K}. 
But for any probability assignment $p$ that is proper for 
\mbox{$\alpha$ \NI $\beta$}, 
\mbox{$p(\alpha \NIm {\bf false})$} is defined and equal to $0$.
Since {\bf K} probabilistically entails 
\mbox{$\alpha$ \NI {\bf false}}
we conclude that there is an $\epsilon > 0$ such that no probability 
assignment that is proper for
{\bf K} and 
\mbox{$\alpha$ \NI $\beta$} 
gives probabilities larger than $1 - \epsilon$ to all assertions of {\bf K}.
Since any probability assignment that is proper for {\bf K} is also proper
for 
\mbox{$\alpha$ \NI $\beta$} 
the conclusion is proved.
\QED
We shall now prove the converse of Lemma~\ref{soundA} in the case 
{\bf K} is finite and probabilistically consistent.
The basic remark is the following. 
Suppose $W$ is a finite (i.e., the set $S$ of states is finite) ranked model.
Let $\epsilon > 0$ be some real number.
We shall describe a probability measure $p_\epsilon$ on $S$.
The first principle that will be used in defining $p_\epsilon$ 
is that all states of the same rank will have equal probabilities.
The second principle is that the weight of the set of all states of rank $n$,
$w_n$ will be such that
\mbox{${w_{n + 1} \over w_n} = \epsilon$}. 
The intuitive meaning of this choice (since $\epsilon$ will approach zero)
is that normal states are more probable than exceptional states.
There is clearly exactly one probability
measure satisfying both principles above, for any given finite ranked model.
The probability measure $p_\epsilon$, defined on states, yields a probability
measure on formulas.
It is clear that a formula $\alpha$ has probability zero under $p_\epsilon$ 
iff $\alpha$ is
inconsistent in $W$, i.e., \mbox{$\alpha$ \NIW {\bf false}}.
Suppose $\alpha$ is consistent.
Let us consider the conditional
probability of $\beta$ given $\alpha$, which is well defined.
If \mbox{$\alpha$ \NIW $\beta$} then this conditional probability
is larger than \mbox{$ 1 - \epsilon - \epsilon^2 - \epsilon^3 - \dots$}
and therefore approaches one when $\epsilon$ approaches zero.
On the other hand, if \mbox{$\alpha$ \notNIW $\beta$}, then this conditional
probability cannot exceed $1 - {1 \over m}$ where $m$ is the number
of states at the rank which is minimal for $\alpha$. It is therefore
bounded away from $1$ when $\epsilon$ approaches $0$.
\begin{lemma}
\label{le:unnamed}
Let a finite probabilistically consistent knowledge base {\bf K} be given 
and suppose 
\mbox{${\cal A} \eqdef \alpha$ \NI $\beta$}
is not preferentially entailed by {\bf K}.
Then, {\bf K} does not probabilistically entail 
${\cal A}$.
\end{lemma}
\proof
Let {\bf K} and ${\cal A}$ be as described in the lemma.
Let ${\cal L'}$ be some logically finite sublanguage of ${\cal L}$ that contains 
$\alpha$, $\beta$ and all propositions appearing in {\bf K}.
Relative to ${\cal L'}$, the hypotheses of the lemma are still true.
By Lemmas~\ref{le:auxrat} and Theorem~\ref{the:new},
there is a rational relation \cE\
that contains \bK\ and \mbox{$\alpha$ \NI $\neg \beta$}
and for which a formula is inconsistent only if it is inconsistent
for \/${\bf K}^p$.
By Theorem~\ref{comthe:rat}, there is a finite ranked model, $W$
that satisfies \mbox{$\alpha$ \NI $\neg \beta$} and all assertions of 
{\bf K} but whose inconsistent formulas are exactly
those of \/${\bf K}^p$. Since \ga\ is not inconsistent for \/${\bf K}^p$,
the model $W$ does not satisfy \ab. Let $W'$ be the model obtained
by extending the labeling function of $W$
to the full language ${\cal L}$ in an arbitrary way.
We shall now apply the construction of $p_\epsilon$ described above on the 
model $W'$.

Using the model $W'$ and a sequence of $\epsilon$'s approaching zero, 
we define 
a sequence of probability measures $p_\epsilon$.
Let us show that all assignments $p_\epsilon$ are proper for {\bf K} and
${\cal A}$. 
If $\gamma \in L'$, the assignment $p_\epsilon$ gives zero probability 
to $\gamma$ iff $\gamma$ is inconsistent in $W$, i.e., inconsistent
for \/${\bf K}^p$.
But {\bf K} is probabilistically consistent and, by Lemma~\ref{133},
$p_\epsilon$ is proper for {\bf K}.
Since $W'$ does not satisfy ${\cal A}$, its antecedent cannot be inconsistent
in $W'$ and $p_\epsilon$ is proper for ${\cal A}$ too.
When $\epsilon$ approaches zero, the conditional probabilities corresponding 
to each assertion of {\bf K} approach $1$ 
and the conditional probability corresponding to
{\cal A} is bounded away from $1$.
\QED
That the result cannot be extended to infinite sets of conditional assertions
follows from Adams' remark that his notion of probabilistic consistency does 
not enjoy
the compactness property and from Corollary~\ref{comp:pref}.
Adams' example~\cite[pages 51--52]{Adams:75} is closely related 
to the construction of Lemma~\ref{le:well}.
The results of Adams presented in this section have been interpreted, 
in particular by~\cite{Pearl:88}, to mean that probabilistic semantics
validate preferential reasoning. We certainly agree. But the results that
will be presented now show, in our opinion, that probabilistic semantics
support the claim that inference procedures should not only be preferential
but also rational. Indeed we show, in Appendix~\ref{appen:nonstandard}, 
that some very natural probabilistic models always define rational relations 
and that, when the language ${\cal L}$ is countable,
all rational relations may be defined by such models. 
Those models are non-standard probability spaces, in the sense of A.~Robinson.
Since no use of those models will be made in the paper, their treatment has
been relegated to an appendix.

\section{The rational closure of a conditional knowledge base}
\label{sec:ratclos}
\subsection{Introduction}
\label{subsec:intro:ratclos}
So far, we have argued for Thesis~\ref{rational} and gathered much
knowledge about rational relations, showing in particular that there
is no obvious way to define a notion of closure satisfying 
Thesis~\ref{rational}.
In this section we shall show that there
is a natural notion of closure (called rational closure)
that satisfies Thesis~\ref{rational}.
We shall study it and prove that it possesses many very elegant
mathematical properties.
We shall, then, evaluate the value of rational closure as an answer to
the question of the title.
In conclusion, we shall propose Thesis~\ref{super}, that claims that any 
satisfactory answer is a superset of rational closure.
In other terms we think that any reasonable system should endorse any
assertion contained in the rational closure, but it may also endorse
some additional assertions.
At present, we do not know of any natural construction satisfying
Thesis~\ref{rational} other than rational closure.

A first possible answer is rejected in~\ref{subsec:perf}.
This result appeared in~\cite{Leh:89}.
The remainder of the paper describes rational closure.
In~\ref{subsec:order} a partial ordering between rational relations is
defined, which captures the notion of a relation being preferable to
(i.e., smaller, less adventurous, more reasonable than) another one. 
The rational closure of a knowledge base is then defined 
in~\ref{subsec:defratclos} as the rational extension of a knowledge base
that is preferable in the ordering defined in Section~\ref{subsec:order}
to all other rational extensions. 
Not every knowledge base has a rational
closure, but in Section~\ref{subsec:proof} it will be shown that 
any admissible (see Definition~\ref{def:acc}) knowledge base has a rational
closure. By Lemma~\ref{co:acc}, then, any finite knowledge base has a 
rational closure. 
We claim that the rational closure of a knowledge base,
when it exists, provides a reasonable answer to the question of the title.
Global properties of the operation
of rational closure are described in~\ref{subsec:ratcum}.
These results, concerning the global behavior of a nonmonotonic inference
operation, are the first of their kind.  
In~\ref{subsec:proof} an algorithmic construction of the rational closure
of an admissible knowledge base is described.
This algorithmic description essentially replaces
and improves upon the proof-theoretic description of~\cite{Leh:89}.
A corrected and generalized model-theoretic construction, 
first described in~\cite{Leh:89} is proposed in~\ref{subsec:modrat}. 
Section~\ref{subsec:comprat} presents an algorithm to compute the 
rational closure of a finite knowledge base and discusses complexity
issues.
Section~\ref{subsec:disc} discusses the appeal of rational closure
and provides some examples.
Section~\ref{sec:con} concludes by considering topics for
further research.
In~\cite{Pearl:90} J.~Pearl proposes his own version of the rational closure
construction that had been described in~\cite{Leh:89}.

\subsection{Perfect extensions}
\label{subsec:perf}
All that has been done so far does not allow us to give a satisfactory answer
to the question of the title.
Let \bK\ be a set of conditional assertions. We would like to define a 
consequence relation \oK, the rational closure of \bK, 
that contains all the conditional assertions that we intuitively expect to
follow from \bK.
At this point the reader should be convinced that \oK\
should be a rational consequence relation that extends \bK.
Any such relation obviously also extends \bKp. 
It seems that we would also like this rational extension of \bK\ 
to be as small as possible.
Unfortunately Theorem~\ref{rat:ent} shows that the intersection of
all rational extensions of \bK\ is exactly \bKp\ and therefore not
in general rational and highly unsuitable as shown in 
Section~\ref{subsec:discpref}.
There is obviously a maximal such extension: the full consequence relation,
(i.e., \mbox{$\alpha$ \NI $\beta$} for all $\alpha$, $\beta$ in ${\cal L}$)
but this is certainly not the one we are looking for.
Can we find out a number of properties that we would like $\overline {\bf K}$ 
to possess, in order to, at least, narrow the field of possibilities? 
We shall look both for {\em local} properties of \oK\ with respect
to \bK\ and for {\em global} properties of the mapping 
\mbox{${\bf K} \mapsto \overline{\bf K}$}.
The sequel will present a proposal for the definition of \oK\
and proofs that it enjoys both local and global (in particular
a strong form of cumulativity) properties.

If \bKp\ happens to be rational, then we probably have no reason to look
further and should take \oK\ to be equal to \bKp.
If \bKp\ is not rational, then there is an assertion \ab\ in \bKp,
and a formula \gc\ such that neither 
\mbox{$\alpha \wedge \gamma$ \NI \gb} nor \mbox{\ga \NI $\neg \gamma$}
are in \bKp.
It seems that the right thing to do, in most such cases, is to introduce 
\mbox{$\alpha \wedge \gamma$ \NI \gb} in \oK.
One may try to require that any assertion in 
\mbox{$\overline {\bf K} - {\bf K}^p$} be of the 
form \mbox{$\alpha \wedge \gamma$ \NI \gb} where \ab\ is in \bKp,
i.e., that any assertion in \mbox{$\overline {\bf K} - {\bf K}^p$} 
have {\em support} in
${\bf K}^p$. It will be shown that this may well be impossible.
Let us encapsulate this idea in definitions.
\begin{definition}
\label{def:supp}
An assertion \mbox{$\alpha$ \NI $\beta$} is said to be supported by
(or in) \/${\bf K}^p$ iff
there is a formula \gc\ such that \mbox{$\alpha \models \gamma$}
and \mbox{$\gamma$ \NI $\beta$} is in
\/${\bf K}^p$.
\end{definition}
\begin{definition}
\label{def:perf}
A rational extension \/${\bf K}'$ of \/{\bf K} is called perfect iff
every assertion of \/${\bf K}'$ is supported by \/${\bf K}^p$.
\end{definition}
We may present the following disappointing result.
\begin{lemma}
\label{disa}
There is a finite conditional knowledge base that has no rational perfect 
extension.
\end{lemma}
\proof
Let ${\cal L}$ be the set of all propositional formulas built out of the set of four
propositional variables: \mbox{$\{a , b , c , d \}$}.
Let $W$ be the preferential model with three states: 
\mbox{$\{ s , t , u \}$}, in which 
\mbox{$s \prec t$} (and this is the only pair in the relation $\prec$)
and $s$ satisfies only $a$, $t$ satisfies only $b$ and $u$ satisfies only
$c$ and $d$.
Let {\bf K} be the set of all conditional assertions satisfied in $W$.
We claim that {\bf K} has no rational perfect extension.
Notice, first, that $W$ satisfies 
\mbox{$ a \vee b$ \NI $\neg b$}. This assertion is therefore in {\bf K}.
Any ranked model satisfying
\mbox{$ a \vee b$ \NI $\neg b$} must satisfy at least one of the following
two assertions:
\mbox{$a \vee c$ \NI $\neg c$} or
\mbox{$b \vee c$ \NI $\neg b$}.
Any rational extension of {\bf K} must therefore contain one of the
two assertions above.
But \mbox{$a \vee c$ \NI $\neg c$} has clearly no support in ${\bf K}^p$
and therefore any perfect rational extension of {\bf K} must contain:
\mbox{$b \vee c$ \NI $\neg b$}.
But $W$ satisfies 
\mbox{$c$ \NI $d$} and any ranked model satisfying both
\mbox{$b \vee c$ \NI $\neg b$} and \mbox{$c$ \NI $d$} must also
satisfy
\mbox{$b \vee c$ \NI $d$}.
Any perfect rational extension of {\bf K} must therefore contain this
last formula but it clearly lacks support in ${\bf K}^p$.
We conclude that {\bf K} has no perfect rational extension. 
\QED
It is therefore reasonable to look for less than perfect extensions.
Let us first examine perfection concerning two special kinds of 
formulas. The following is easily proved.
\begin{lemma}
\label{le:perf}
An assertion of the form \mbox{$\alpha$ \NI {\bf false}} is supported
by \/${\bf K}^p$ iff it is in \/${\bf K}^p$.
An assertion of the form \mbox{{\bf true} \NI $\alpha$} is supported
by \/${\bf K}^p$ iff it is in \/${\bf K}^p$.
\end{lemma}
We shall propose a construction of \oK\ such that \oK\ does not contain
any formula of the
form \mbox{$\alpha$ \NI {\bf false}} or of the form 
\mbox{{\bf true} \NI $\alpha$}
that is not in \bKp.

\subsection{Ordering rational relations}
\label{subsec:order}
In this section we shall define a strict partial ordering between 
rational relations. This ordering captures the notion
of a relation being preferable to, i.e., less adventurous than another one.
An intuitive explanation will be given immediately after the definition.
For the rest of this section we shall write 
{\em \mbox{$\alpha < \beta$} for ( or in) K} to mean {\em the assertion 
\mbox{$\alpha \vee \beta$ \NI $\neg \beta$} is in $K$}.
We shall write \mbox{$\alpha \leq \beta$} for ( or in) $K$ when it
is {\em not} the case that \mbox{\gb $<$\ga} in $K$. 
\begin{definition}
\label{def:order}
Let $K_0$ and $K_1$ be two rational consequence relations.
We shall say that $K_0$ is preferable to $K_1$ and write
$K_0 \prec K_1$ iff: 
\begin{enumerate}
\item \label{<1}
there exists an assertion \ab\ in \mbox{$K_1 - K_0$} such that for all \gc\ 
such that
\mbox{$\gamma < \alpha$} for $K_0$, and for all \gd\ such that
\mbox{\gc \NI \gd} is in $K_0$, we also have \mbox{\gc \NI \gd} in $K_1$,
and
\item \label{<2}
for any \gc, \gd\, if \mbox{\gc \NI \gd} is in \mbox{$K_0 - K_1$}  there is 
an assertion \mbox{$\rho \NIm \eta$} in \mbox{$K_1 - K_0$} such that
 \mbox{$\rho < \gamma $} for $K_1$.
\end{enumerate}
\end{definition}
The intuitive explanation behind Definition~\ref{def:order}
is the following. Suppose two agents,
who agree on a common knowledge base, 
are discussing the respective merits of two rational relations
$K_0$ and $K_1$.
A typical attack would be: {\em your relation contains an assertion,
\ab, 
that mine does not contain} (and therefore contains unsupported assertions).
A possible defense against such an attack could be:
{\em yes, but your relation contains an assertion 
\mbox{\gc \NI \gd} that mine does not, and you yourself think that \gc\ 
refers to a situation that is more usual than the one refered to by \ga}.
Such a defense must be accepted as valid.
Definition~\ref{def:order} exactly says that the proponent of $K_0$
has an attack that the proponent of $K_1$ cannot defend against
(this is part~\ref{<1}) but that he (i.e., the proponent of $K_0$) may
find a defense against any attack from the proponent of $K_1$
(this is part~\ref{<2} of the definition).

\begin{lemma}
\label{le:order}
The relation $\prec$ between rational consequence relations is
irreflexive and transitive.
\end{lemma}
\proof
Irreflexivity follows immediately from Condition~\ref{<1}.
For transitivity, let us suppose that $K_0 \prec K_1$, with \ab\ the witness
promised by Condition~\ref{<1} and that \mbox{$K_1 \prec K_2$} 
with \cd\ as a witness. 
Our first step will be to show that there exists an assertion \ef\ in
$K_2 - K_0$ such that \mbox{$\varepsilon \leq \alpha$} in $K_0$ and
\mbox{$\varepsilon \leq \gamma$} in $K_1$. We shall have to consider 
many different cases.
\begin{enumerate}
\item Suppose \mbox{$\gamma < \alpha$} in $K_2$.
\begin{enumerate}
\item If \mbox{$\gamma < \alpha$} is not in $K_0$, then 
\mbox{$\gamma < \alpha$} is a suitable \ef.
\item If \mbox{$\gamma < \alpha$} is in $K_0$, then \cd\ is a suitable
\ef, since if it were in $K_0$ it would be in $K_1$.
\end{enumerate}
\item Suppose therefore that \mbox{$\gamma < \alpha$} is not in $K_2$, 
i.e., for $K_2$, \mbox{$\alpha \leq \gamma$}.
\begin{enumerate}
\item If \mbox{$\gamma < \alpha$} is in $K_1$, then it is in $K_1 - K_2$
and there is an assertion \xe\ in $K_2 - K_1$ such that 
\mbox{$\xi < \gamma \vee \alpha$} in $K_2$.
\begin{enumerate}
\item If \mbox{$\xi < \gamma \vee \alpha$} is not in $K_0$, then
it is a suitable \ef.
\item If \mbox{$\xi < \gamma \vee \alpha$} is in $K_0$, then 
\mbox{$\xi < \alpha$} in $K_0$ and we have both that
\mbox{$\xi < \gamma \vee \alpha$} in $K_1$ and that \xe\ cannot be in
$K_0$, otherwise it would be in $K_1$. 
We conclude that \xe\ is a suitable \ef.
\end{enumerate}
\item Suppose therefore that \mbox{$\gamma < \alpha$} is not in $K_1$,
i.e., for $K_1$, like for $K_2$, \mbox{$\alpha \leq \gamma$}.
\begin{enumerate}
\item If \ab\ is in $K_2$, then it is a suitable \ef.
\item If \ab\ is not in $K_2$, then it is in $K_1 - K_2$ and there is
a \xe\ in $K_2 - K_1$ such that \mbox{$\xi < \alpha$} in $K_2$.
\begin{enumerate}
\item If \mbox{$\xi < \alpha$} is not in $K_0$, then it is a suitable 
\ef\ since, in $K_1$, \mbox{$\xi \vee \alpha \leq \alpha \leq \gamma$}.
\item If \mbox{$\xi < \alpha$} is in $K_0$ then \xe\ is a suitable \ef\
since \xe\ cannot be in $K_0$, otherwise it would be in $K_1$.
\end{enumerate}
\end{enumerate}
\end{enumerate}
\end{enumerate}

We have now proved the existence of an assertion \ef\ with the desired 
properties.
Let us proceed to the proof that $K_0 \prec K_2$. For property~\ref{<1},
we claim that \ef\ provides a suitable witness. It is indeed in $K_2 - K_0$
by construction. Suppose now that \mbox{$\xi < \varepsilon$} in $K_0$.
Then \mbox{$\xi < \alpha$} in $K_0$ and therefore 
\mbox{$\xi < \varepsilon$} in $K_1$.
Therefore \mbox{$\xi < \gamma$} in $K_1$. If \xe\ is in $K_0$, then it must
be in $K_1$ since \mbox{$\xi < \alpha$} in $K_0$ and also in $K_2$ since 
\mbox{$\xi < \gamma$} in $K_1$. This concludes the verification 
of Condition~\ref{<1}.

For Condition~\ref{<2}, suppose that \pht\ is in $K_0 - K_2$.
We have to find a \xe\ in $K_2 - K_0$ such that \mbox{$\xi < \varphi$}
in $K_2$. We shall consider a number of different cases.
\begin{enumerate}
\item If \mbox{$\varepsilon < \varphi$} in $K_2$, then \ef\ is a suitable
\xe.
\item Suppose then that \mbox{$\varepsilon < \varphi$} is not in $K_2$, i.e., 
\mbox{$\varphi \leq \varepsilon$} for $K_2$.
\begin{enumerate}
\item Suppose, first that \mbox{$\varepsilon < \varphi$} is in $K_1$,
therefore in $K_1 - K_2$. There is then an assertion \rt\ in $K_2 - K_1$
such that \mbox{$\rho < \varepsilon \vee \varphi$} in $K_2$.
\begin{enumerate}
\item If \mbox{$\rho < \varepsilon \vee \varphi$} is in $K_0$, then
\mbox{$\rho < \alpha$} in $K_0$ and we conclude that \rt\ is not in $K_0$,
otherwise it would be in $K_1$. We conclude that \rt\ is a suitable
\xe.
\item If \mbox{$\rho < \varepsilon \vee \varphi$} is not in $K_0$, then it is
a suitable \xe.
\end{enumerate}
\item Suppose, then, that \mbox{$\varepsilon < \varphi$} is not in $K_1$,
i.e., for $K_1$, \mbox{$\varphi \leq \varepsilon$}.
\begin{enumerate}
\item If \mbox{$\varepsilon < \varphi$} is in $K_0$, then it is in $K_0 - K_1$.
Therefore there is an assertion \rt\ in $K_1 - K_0$ such that
\mbox{$\rho < \varepsilon \vee \varphi$} in $K_1$. But then 
\mbox{$\rho < \gamma$} in $K_1$ and we conclude that \rt\ is in $K_2$
and that \mbox{$\rho < \varphi$} in $K_2$. The assertion \rt\ is a suitable
\xe.
\item Suppose, then that, on the contrary, 
\mbox{$\varepsilon < \varphi$} is not in $K_0$, 
i.e., \mbox{$\varphi \leq \varepsilon$}
in $K_0$, as in $K_1$ and $K_2$.
\begin{enumerate}
\item Suppose first that \pht\ is in $K_1$, therefore in $K_1 - K_2$.
Then, there is an assertion \rt\ in $K_2 - K_1$, such that 
\mbox{$\rho < \varphi$} in $K_2$.
There are two cases.
If \mbox{$\rho < \varphi$} is in $K_0$, then \mbox{$\rho < \varepsilon$}
in $K_0$, and \rt\ is not in $K_0$, otherwise it would be in $K_1$,
since \mbox{$\rho < \varepsilon \leq \alpha$} in $K_0$.
The assertion \rt\ is a suitable \xe.
If, on the other hand 
\mbox{$\rho < \varphi$} is not in $K_0$, then it is in $K_2 - K_0$,
and it is a suitable \xe.
\item Suppose now that \pht\ is not in $K_1$, therefore in \mbox{$K_0 - K_1$}.
There is an assertion \rt\ in \mbox{$K_1 - K_0$}, such that 
\mbox{$\rho < \varphi$} in $K_1$. But \mbox{$\rho < \varphi \leq \varepsilon 
\leq \gamma$} in $K_1$ and since \rt\ is in $K_1$ it must be in $K_2$.
Also, since \mbox{$\rho < \varphi$} is in $K_1$, it must be in $K_2$.
We see that \rt\ is a suitable \xe.
\end{enumerate}
\end{enumerate}
\end{enumerate}
\end{enumerate}
\QED

\subsection{Definition of rational closure}
\label{subsec:defratclos}
We may now define the rational closure of a knowledge base.
\begin{definition}
\label{def:ratclos}
Let \bK\ be an arbitrary knowledge base. 
If there is a rational extension \oK\ of \bK\ that is preferable to all other
rational extensions of \bK, then \oK\ will be called the rational closure
of \bK.
\end{definition}
Notice first that the rational closure of a knowledge base is unique,
if it exists, since {\em preference} is a partial ordering.
Notice then that there are knowledge bases that do not
have a rational closure. Example~\ref{ex:noclos} will show this.
In Section~\ref{subsec:proof} we shall show that admissible  
knowledge bases, including all finite knowledge bases, have a rational
closure.

\begin{example}
\label{ex:noclos}
{\rm
Let ${\cal L}$ be the propositional calculus built upon the variables
$p_n$ where $n$ is an arbitrary {\em integer} (i.e., positive or negative).
Let $N$ be the knowledge base that contains all assertions of the form 
\mbox{$p_n \NIm p_{n + 2}$} and of the form 
\mbox{$p_n \NIm \neg p_{n - 2}$} for all integers $n$.
We shall show that $N$ has no rational closure.
}
\end{example}

We shall first prove a lemma about invariance of the operation of 
rational closure under renaming of the proposional variables.
This lemma is of independent interest.

\begin{definition}
\label{def:rename}
\begin{enumerate}
\item A renaming of the propositional calculus ${\cal L}$ is a bijection of the
 propositional variables.
\item Let $f$ be a renaming of ${\cal L}$.
The formula obtained from \ga\ by substituting $f(p)$
for the propositional variable $p$ will be denoted by $f(\ga)$.
\item Let $f$ be as above and \ab\ a conditional assertion.
The assertion $f(\ga)\NIm f(\gb)$ will be denoted by $f(\alpha \NIm \beta)$.
\item Let $f$ be as above and $K$ a consequence relation. The relation 
$f(K)$ will be defined by \mbox{$f(K) = \{f(\ab)\mid\ab\in K\}$}.
\end{enumerate}
\end{definition}

\begin{lemma}[Invariance under renaming]
\label{lemma:inv-renaming}
Let $f$ be a renaming of ${\cal L}$.
\begin{enumerate}
\item
Let $K_0$ and $K_1$ be rational consequence relations.
Then $K_0 \prec K_1$ iff $f(K_0) \prec f(K_1)$.
\item Let $K$ be a consequence relation and $\overline{K}$ 
its rational closure ,then
$f(\overline{K})$ is the rational closure of $f(K)$
\item Let $K$ be a consequence relation which is invariant under $f$,
namely $f(K)=K$,then
its rational closure (if it exists) is invariant under $f$.
\end{enumerate}
\end{lemma}

\proof 
The proof is immediate from the definitions,
noting that $f$ is also a bijection
of the set of all consequence relations.
\QED

\begin{lemma}
The knowledge base $N$ defined above has no rational closure.
\end{lemma}
\proof 
We shall reason by contradiction.
Suppose $R$ is the rational closure of $N$.
From Lemma~\ref{le:cperf} in the sequel (the proof of which does not
depend on the present lemma), we know that there is no assertion
of the form $\gafalse$ in $R$ that is not in $N^p$.
Using a construction very similar to the one used in the proof of
Lemma~\ref{le:well}, one may build, for any integer $n$, a preferential
model of $N$, containing a top state that satisfies $p_{n}$.
Therefore, for any $n$, $p_{n}$ is consistent for $R$.

Remember that \mbox{$a < b$} is the assertion \mbox{$a \vee b$ \NI $\neg b$}.
We shall write \mbox{$a < b$} to mean that \mbox{$a \vee b$ \NI $\neg b$} is in
$R$.
If follows from results of Section~\ref{subsec:rank} that,
on formulas that are consistent for $R$, the relation $<$ is a strict
modular ordering.
Notice, also, that, for any $n$, the assertion 
\mbox{$p_{n+2} < p_{n}$}
belongs to $N^p$, since both \mbox{$p_{n}$ \NI $p_{n+2}$} and
\mbox{$p_{n+2}$ \NI $\neg p_{n}$} are in $N^p$.
Therefore \mbox{$p_{n+2} < p_{n}$}. 
There are, in $R$, two infinite (in both directions) chains (for $<$), 
one containing the variables of odd index, 
the other one containing those of even index.
Since $<$ is modular, we may consider only four cases:
\begin{enumerate}
\item
\label{exo1}
For every even $n$ and odd $k$, \mbox{$p_n>p_k$}.
Let $f$ be the renaming of ${\cal L}$ given by \mbox{$f(m)=m+1$}.
Clearly \mbox{$f(N)=N$}.
Hence, by Lemma ~\ref{lemma:inv-renaming} we must have 
\mbox{$f(R)=R$}.
But this last statement implies \mbox{$p_n>p_k$} for even $n$ 
and odd $k$, and therefore implies that all $p_n$'s are inconsistent
for $R$. A contradiction.
\item
\label{exo2}
For every even $n$ and odd $k$, \mbox{$p_n<p_k$}.
The argument is exactly as in case~\ref{exo1},
systematically interchanging `odd' and `even'.
\item
\label{exo3}
There is an odd $k$ ,and there are even $m$ and $n$ such that 
\mbox{$p_n<p_k<p_m$}.
In this case, define a renaming $f$ by $f(l)=l$ for odd $l$ and $f(l)=l+m-n$
for even $l$.
The contradiction is as above by noting that $f$ transforms
\mbox{$ p_n < p_k$} into \mbox{$p_m < p_k$}.
\item
\label{exo4}
None of the above is true. In such a case one may see that there must
exist an even $m$ ,and odd $i$ and $j$ such that 
\mbox{$p_i<p_m<p_j$}. The argument is exactly as in case~\ref{exo3},
systematically interchanging `odd' and `even'.
\end{enumerate}
\QED
 
\subsection{Global properties of the operation of rational closure}
\label{subsec:ratcum}
First, we show that rational closure possesses a {\em loop} property
analogous to the property discussed in~\cite[Section 4]{KLMAI:89}.
This is a powerful property that one is happy to have.
\begin{lemma}[Loop property]
\label{le:loop}
Let $K_i$ for $i = 0 , \ldots , n-1$ be knowledge bases such that,
for any $i$, \mbox{$K_{i +1} \subseteq \overline{K_{i}}$}, where addition 
is understood modulo $n$. Then for any $i , j$, one has
\mbox{$\overline{K_{i}} = \overline{K_{j}}$}.
\end{lemma}
\proof
Let $K \preceq K'$ mean that either $K \prec K'$ or $K = K'$. 
Since $\overline{K_{i}}$ is a rational extension of $K_{i+1}$,
we have \mbox{$\overline{K_{i+1}} \preceq \overline{K_{i}}$},
for all $i$'s (modulo $n$).
We conclude that the rational closures of all the ${\bf K}_i$'s are equal.
\QED
The following property of {\em reciprocity} is the special case $n = 2$.
\begin{corollary}
\label{co:rec}
If \mbox{$X \subseteq \overline{Y}$} and 
\mbox{$Y \subseteq \overline{X}$}, then 
\mbox{$\overline{X} = \overline{Y}$}.
\end{corollary}
The following property of {\em cumulativity} is equivalent to reciprocity
in the presence of inclusion (i.e., \mbox{$K \subseteq \overline{K}$}).

\begin{corollary}
\label{co:cum}
If \mbox{$X \subseteq Y \subseteq \overline{X}$} then
\mbox{$\overline{X} = \overline{Y}$}.
\end{corollary}
The meaning of Corollary~\ref{co:cum} is that one may add to a knowledge
base anything that is in its rational closure without changing this
closure.
We may now show that, in two different respects, rational closure is
close to being perfect.

\begin{lemma}
\label{le:tperf}
The consequence relation \oK, if it exists, contains an assertion of the form 
\mbox{{\bf true} \NI \ga} only if this assertion is in \bKp.
\end{lemma}
\proof
Suppose an assertion of the form above is in \oK. 
We shall show that it must be in any rational extension of \bK\
and will conclude by Theorem~\ref{rat:ent}.
Suppose $K'$ is a rational extension of \bK\ and \mbox{{\bf true} \NI \ga}
is in \mbox{$\overline{\bf K} - K'$}. 
Since \oK $\prec K'$, we know there is an assertion 
\mbox{$\gamma \NIm \delta$} in $K' - \overline{\bf K}$ such that
\mbox{$\gamma <$ {\bf true}} in $K'$. But this means
\mbox{{\bf true} \NI {\bf false}} is in $K'$, and 
contradicts the fact that \mbox{{\bf true} \NI \ga} is not in $K'$.
\QED

\begin{lemma}
\label{le:cperf}
The consequence relation \oK, if it exists, contains an assertion of the form 
\mbox{\ga \NI {\bf false}} only if this assertion is in \bKp.
\end{lemma}
\proof
Let $V$ and $T$ be preferential models defining the relations \oK\ and
\bKp\ respectively. Such models exist by Theorem~\ref{compth:pref}.
Let $U$ be the model obtained by putting $T$ on top of $V$, i.e.,
every state of $V$ is less than every state of $T$. One easily sees
that $U$ satisfies the smoothness property and is therefore a preferential
model. It defines a preferential relation $S$.
An assertion of the form \mbox{\ga \NI {\bf false}} is in $S$ only if it
is in \bKp, since $T$ is a submodel of $U$. If \ga\ is not inconsistent
in \oK\, then for any \gb, \ab\ is in $S$ iff it is in \oK.
By Lemma~\ref{le:auxrat}, there is a rational extension $R$ of $S$
with the same set of inconsistent formulas. If one looks at the
construction described in the proof of this lemma, one sees that
it will add to $S$ only assertions with antecedent inconsistent in \oK.
Therefore, if \ga\ is not inconsistent
in \oK, for any \gb, \ab\ is in $R$ iff it is in \oK.
Now, $R$ is a rational extension of \bK. 
If $R$ is equal to \oK, we are through.
Suppose not. Then we have \mbox{$\overline{\bf K} \prec R$}.
Suppose now that \mbox{\ga \NI {\bf false}} is in 
\mbox{$\overline{\bf K} - R$}. There must be an assertion \cd\ in
\mbox{$R - \overline{\bf K}$} such that \mbox{$\gamma < \alpha$} in $R$.
But \cd\ in \mbox{$R - \overline{\bf K}$} implies that
\mbox{\gc \NI {\bf false}} is in \oK\ and \cd\ is in \oK. A contradiction.
\QED

\subsection{Admissible knowledge bases and their rational closure}
\label{subsec:proof}

In this section, we show that an admissible (see Definition~\ref{def:acc})
knowledge base has a rational 
closure and that this rational closure may be defined in terms of the ranks
of the formulas, as defined in Section~\ref{subsec:ranking}.
This provides a useful and elegant characterization of the rational closure
of an admissible knowledge base.

\begin{theorem}
\label{the:ratclos}
Let \bK\ be an admissible conditional knowledge base. 
The rational closure \oK\ of \bK\ exists and is 
the set $S$ of all assertions \ab\ such that either
\begin{enumerate}
\item the rank of \ga\ is strictly less than the rank of 
\mbox{$\alpha \wedge \neg \beta$} (this includes the case 
\ga\ has a rank and \mbox{$\alpha \wedge \neg \beta$} has none), or
\item \ga\ has no rank (In this case \mbox{$\alpha \wedge \neg \beta$}
has no rank either).
\end{enumerate}
\end{theorem}

\proof
Suppose indeed that every formula consistent with \bKp\ has a rank.
We have many things to check. First let us prove that $S$ contains \bK.
If \ab\ is in \bK\ and \ga\ has rank $\tau$, then $C_{\tau}$ contains
\ab\ and entails \mbox{{\bf true} \NI $\alpha \ra \beta$}.
Therefore \mbox{$\alpha \wedge \neg \beta$} is exceptional for
$C_{\tau}$, and has rank strictly larger than $\tau$.

We should now check that $S$ is rational.
For {\bf Left Logical Equivalence}, {\bf Right Weakening} and
{\bf Reflexivity} the proof is easy.
For {\bf Cautious Monotonicity}, notice that if
\ab\ is in $S$, then \ga\ and \mbox{$\alpha \wedge \beta$} have the same rank.
For {\bf And} and {\bf Or}, 
notice that the rank of a disjunction is the smaller
of the ranks of its components.
For {\bf Rational Monotonicity}, notice that if
\mbox{$\alpha \NIm \neg \beta$} is not in $S$, then \ga\ and 
\mbox{$\alpha \wedge \beta$} have the same rank.

We must now check that if $R$ is a rational extension of \bK\ that is
different from $S$ then \mbox{$S \prec R$}.
Let $R$ be such an extension. 
We shall first show that $S$ and $R$
must agree on all assertions whose antecedents have no rank 
(the notion of rank is always defined by reference to \bK).
Indeed, by construction, any such assertion is in
$S$, and it is preferentially entailed by \bK\ since \bK\ is admissible.
It is therefore in $R$.
We conclude that $S$ and $R$ must differ for some assertion that
has rank. 
Let $\tau$ be the smallest rank at which $S$ and $R$ differ,
i.e., the smallest rank of an \ga\ such that there is a \gb\ such that
\mbox{$\alpha \NIm \beta \in ( S - R ) \cup ( R - S )$}.
We have two cases to consider, either there is a formula \ga\ 
of rank greater or equal to $\tau$ such that,
for all formulas \gb\ of rank greater or equal to $\tau$, 
\mbox{$\alpha \leq \beta$}
in $R$, or there is no such formula.

Suppose there is such an \ga.
Our first claim is that, for any
\gb\ of rank greater than $\tau$, 
the assertion \mbox{$\alpha < \beta$} is in $R$.
Consider indeed a ranked model $W$ that defines $R$.
Let $W''$ be the supermodel
obtained from $W$ by adding to $W$, at each level $l$ a state labeled
with world $w$ for all worlds $w$ that label a state of rank less than
$l$ in $W$. It is clear that $W''$ is ranked and defines the same
relation as $W$, and, in $W''$, every label that appears 
at some level $l$ also appears at all greater levels.
Let $W'$ be the submodel of $W''$ that contains all those states of
level (rank in $W''$) greater or equal to the minimal level $l$ 
at which some state
satisfies \ga. It clearly satisfies the smoothness property
(for this we needed to go through the construction of $W''$).
Since a formula is satisfied in $W''$ at some level less than $l$ iff
it is of rank less than $\tau$, no antecedent of
an assertion of $C_{\tau}$ is satisfied at any level less than $l$. 
But $W''$ is a model of \bK\ and therefore $W'$ 
is a model of $C_{\tau}$.
But $C_{\tau}$ preferentially entails 
\mbox{{\bf true} \NI $\neg \beta$}. The model $W'$ therefore
satisfies \mbox{{\bf true} \NI $\neg \beta$} but not 
\mbox{{\bf true} \NI $\neg \alpha$}. It therefore also
satisfies \mbox{$\alpha < \beta$}. But the antecedent of this last assertion
has rank greater or equal to $\tau$, and therefore no state of $W''$ that is
not in $W'$ satisfies it. Therefore \mbox{$\alpha < \beta$} is satisfied by
$W''$ and is an element of $R$.
satisfies
Our second claim is that there is an assertion \cd\ 
in \mbox{$R - S$}, such that \gc\ is of rank $\tau$ and 
\mbox{$\gamma \leq \alpha$} in $R$.
We consider two cases.
\begin{enumerate}
\item
There is an assertion \xe\ in \mbox{$S - R$} with $\xi$ of rank $\tau$. 
Then \mbox{$\xi \wedge \neg \eta$} has
rank greater than $\tau$, and by our first claim,
\mbox{$\alpha < \xi \wedge \neg \eta$} is in $R$.
But \xe\ is not in $R$, and therefore we must have
\mbox{$\alpha < \xi$} for $R$.
This last assertion is not in $S$ since both
\ga\ and $\xi$ have rank $\tau$.
The assertion \mbox{$\alpha < \xi$} is a suitable \cd.
\item
There is an assertion \xe\ in \mbox{$R - S$} with $\xi$ of rank $\tau$.
If \mbox{$\xi \leq \alpha$} in $R$, then \xe\ is a suitable \cd.
Suppose, then, that \mbox{$\alpha < \xi$} in $R$. Since $\xi$ has the same
rank as \ga, \mbox{$\alpha < \xi$} is in \mbox{$R - S$} and a suitable
\cd.
\end{enumerate}

We may now conclude that \mbox{$S \prec R$}.
The assertion \cd\ fulfills the requirements of Condition~\ref{<1}
of Definition~\ref{def:order}, since \gc\ has rank $\tau$.
For Condition~\ref{<2}, suppose \xe\ is in \mbox{$S - R$},
then $\xi$ must be of rank greater or equal to $\tau$
and \mbox{$\xi \wedge \neg \eta$} is of rank greater than $\tau$.
By our first remark we conclude that 
\mbox{$\alpha < \xi \wedge \neg \eta$} for $R$.
It is a matter of elementary properties of rational relations to check that
if \mbox{$\alpha < \xi \wedge \neg \eta$} is in $R$, but \xe\ is not,
then \mbox{$\alpha < \xi$} for $R$. Since \mbox{$\gamma \leq \alpha$} in $R$,
we conclude that \mbox{$\gamma < \xi$} for $R$.

Suppose now that there is no such \ga.
Take any formula \gc\ of rank $\tau$. There is a formula \gd\ of
rank greater or equal to $\tau$ such that \mbox{$\delta < \gamma$} for $R$.
But this assertion is then in \mbox{$R - S$}. It satisfies Condition~\ref{<1}
of Definition~\ref{def:order}, since its antecedent has rank $\tau$.
Suppose now \pht\ is in \mbox{$S - R$}. Then $\varphi$ is of rank at least
$\tau$. If it is of rank $\tau$, there is a formula $\pi$ of 
rank at least $\tau$ such that \mbox{$\pi < \varphi$} is in $R$, 
but not in $S$ and
this provides the witness requested by Condition~\ref{<2}.
If it is of rank greater than $\tau$, then the assertion 
\mbox{$\delta < \gamma$} defined just above will do.
\QED

\subsection{A model-theoretic description of rational closure}
\label{subsec:modrat}
We shall describe here a model-theoretic construction that transforms
a preferential model $W$ into a ranked model $W'$ by letting all states 
of $W$ sink as low as they can respecting the order of $W$,
i.e., ranks the states of $W$ by their height in $W$.
We shall show that, under certain conditions, the model $W'$ defines
the rational closure of the relation defined by $W$.
This construction is clearly interesting only when the model 
$W$ is well-founded.
We know that, in this case, the relation defined by $W$ indeed possesses
a rational closure (Theorem~\ref{the:ratclos} and Lemma~\ref{co:acc}).
It would have been pleasant to be able to prove the validity of such
a construction on an arbitrary well-founded preferential model.
Unfortunately we are not able to show this in general, but need to
suppose, in addition, that the preferential relation defined by $W$ is 
well-founded (see Definition~\ref{def:wellf}).
This is quite a severe restriction, since we have seen at the end of 
Section~\ref{subsec:prefmod} that
finitely-generated relations on arbitrary languages ${\cal L}$
are not always well-founded.
When the language ${\cal L}$ is logically finite, we know all
preferential relations are well-founded.
Given a well-founded preferential relation, the construction may be applied 
to any of its well-founded models.

Let $P$ be a well-founded preferential relation and 
\mbox{$W = \langle S, l, \prec \rangle$} any well-founded preferential
model that defines $P$.
We shall define, for any ordinal $\tau$, two sets of states: $U_{\tau}$
and $V_{\tau}$.
Those sets satisfy, for any $\tau$,
\mbox{$U_{\tau} \subseteq V_{\tau} \subseteq U_{\tau + 1}$}.
The set $U_{\tau}$ contains, in addition to the elements of previous
$V$'s, the states that are minimal among those states not previously 
added.
The set $V_{\tau}$ contains, in addition to the states of $U_{\tau}$,
all states that satisfy only formulas already satisfied by states
previously considered.
\begin{displaymath}
U_{\tau} \eqdef \bigcup_{\rho < \tau}{V_{\rho}} 
\end{displaymath}
\begin{equation}
\label{eq:Utau}
\cup \{ s \in S \mid 
\forall t \in S {\rm \ such \  that\ } t \prec s, {\rm there \  is \  a \ }
\rho < \tau {\rm \ such \  that \ } t \in V_{\rho} \}
\end{equation}
\begin{equation}
\label{eq:Vtau}
V_{\tau} \eqdef \{ s \in S \mid \forall \alpha \in {\cal L} 
{\rm \ such \ that \ }
s \EM \alpha, \exists t \in U_{\tau} {\rm \ such \  that \ }
t \EM \alpha\}
\end{equation}
Since the model $W$ is well-founded, every state \mbox{$s \in S$} is in
some $V_{\tau}$.
Let the height of a state \mbox{$s \in S$} (in $W$) be the least ordinal $\tau$
for which \mbox{$s \in V_{\tau}$}.
We shall now show that there is a close relationship between
the rank of a formula \ga\ in $P$ (see definition following 
Definition~\ref{def:exc}) and the height in $W$ of the states
that satisfy \ga.
For any ordinal $\tau$,
we shall denote by $W_\tau$ the substructure of $W$ consisting of all 
states of height larger or equal to $\tau$. 
Notice that, since $W$ is well-founded, $W_\tau$ is a preferential model.
Notice also that all elements of 
\mbox{$U_{\tau} - \bigcup_{\rho < \tau}{V_{\rho}}$} are minimal elements of 
$W_{\tau}$.

\begin{lemma}
\label{le:Sch}
Let $\tau$ be an ordinal.
Let \ga\ be a formula of rank at least $\tau$ and \gb\ be any formula.
\begin{enumerate}
\item
No state of height less than $\tau$ satisfies \ga.
\item
The model $W_\tau$ satisfies \ab\
iff \ab\ is preferentially entailed by $C_{\tau}$.
\end{enumerate}
In particular, if \ga\ has no rank, no state in $S$ satisfies \ga.
\end{lemma}
\proof
It proceeds by simultaneous ordinal induction on $\tau$.
Suppose both claims have been proved for all ordinals \mbox{$\rho < \tau$}.
Let us prove our first claim.
Since \ga\ has rank at least $\tau$, for any $\rho$, \mbox{$\rho < \tau$},
$C_{\rho}$ preferentially entails \mbox{{\bf true} \NI $\neg \alpha$}.
By the induction hypothesis (item 2), $W_{\rho}$ 
satisfies \mbox{{\bf true} \NI $\neg \alpha$}.
Therefore no state of \mbox{$U_{\rho} - \bigcup_{\sigma < \rho}{V_{\sigma}}$}
satisfies \ga.
If there were a state $s$ of height \mbox{$\rho < \tau$} satisfying \ga, 
there would be a state $t$ of 
\mbox{$U_{\rho} - \bigcup_{\sigma < \rho}{V_{\sigma}}$} satisfying \ga.
We conclude that no state of height less than $\tau$ satisfies \ga.

For the second claim, by Lemma~\ref{le:funda}, \ab\ is preferentially entailed
by $C_{0}$ (i.e., in $P$, i.e., satisfied by $W$) 
iff it is preferentially entailed by $C_{\tau}$.
By the first claim, \ab\ is satisfied by $W$ iff it is satisfied by
$W_{\tau}$.
\QED

\begin{lemma}
\label{le:ra}
A formula \ga\ has rank $\tau$ in $P$ iff 
there is a state \mbox{$s \in S$} of height $\tau$
that satisfies \ga\  and there is no such state of height less than $\tau$.
\end{lemma}
\proof
We shall prove the {\em only if} part. The {\em if} part is then obvious.
First, remark that if \NI is a preferential relation that contains
the assertion
\mbox{\ga $\vee$\gb \NI $\neg$\ga} , then it contains the assertion
\mbox{{\bf true} \NI $\neg$\ga}.
This is easily shown by preferential reasoning.
Suppose now that \ga\ has rank $\tau$.
Lemma~\ref{le:Sch} shows that no state of height less than $\tau$ 
satisfies \ga.
We must show that there is a state of height $\tau$ satisfying \ga.
Let \gb\ be any formula of rank larger or equal to $\tau$ that is minimal
with respect to $<$ among those formulas. There is such a formula since the set
is not empty (\ga\ is there) and $<$ is well-founded.
Since \ga\ is not exceptional for $C_{\tau}$, the assertion
\mbox{{\bf true} \NI $\neg$\ga} is not preferentially entailed by $C_{\tau}$
and therefore the assertion \mbox{\ga $\vee$\gb \NI $\neg$\ga} is
not preferentially entailed by $C_{\tau}$.
But \mbox{\ga $\vee$\gb} has rank $\tau$ and, by Lemma~\ref{le:Sch},
$W_{\tau}$ does not satisfy \mbox{\ga $\vee$\gb \NI $\neg$\ga}.
There is, therefore, in $W_{\tau}$ a state $s$ satisfying \ga\ such that
no state $t$ in $W_{\tau}$, $t \prec s$, satisfies \gb.
We shall show that $s$ is minimal in $W_{\tau}$ and has therefore height 
$\tau$.
Suppose $s$ is not minimal in $W_{\tau}$. 
There would be a state $t$ minimal in  $W_{\tau}$ such that $t \prec s$.
This state $t$ has height $\tau$ and, by construction, it satisfies some
formula $\beta'$ that is not satisfied at any smaller height.
By Lemma~\ref{le:Sch}, $\beta'$ has rank larger or equal to $\tau$,
and the formula \mbox{$\beta \vee \beta'$} has rank larger or equal to
$\tau$.
Since \mbox{$\beta \vee \beta' \leq \beta$}, the minimality of \gb\ 
implies that \mbox{$\beta \leq \beta \vee \beta'$}.
In other terms, \mbox{$\beta \vee \beta' \NI \beta$}.
But the state $t$, in $W$, satisfies $\beta'$ and is minimal among
states satisfying \mbox{$\beta \vee \beta'$}.
Therefore $t$ satisfies \gb. A contradiction.
\QED
Lemma~\ref{le:ra} shows that, given a well-founded preferential relation
(resp. a finite knowledge base), and a well-founded preferential model $W$
for it (resp. for its preferential closure), one may build a ranked model
for its rational closure by ranking the states of $W$ by their depth.

\subsection{Computing rational closure}
\label{subsec:comprat}

We shall now provide an algorithm for deciding whether an assertion
is in the rational closure of a finite knowledge base.
The notation $E(C)$ has been defined following Definition~\ref{def:exc}.
Lemma~\ref{co:acc} and Theorem~\ref{the:ratclos} show that, 
given a finite knowledge base \bK\ and an assertion \ab\,
the following algorithm is adequate.
\begin{quote}
$C = \bK$;

while \ga\ is exceptional for $C$ and $E(C) \neq C$, $C := E( C )$;

if $\alpha \wedge \neg \beta$ is exceptional for $C$ 
			then answer yes
			else answer no.
\end{quote}

The only thing left for us to implement is checking whether a formula
is exceptional for a given finite knowledge base.
The next lemma shows this is easily done.

\begin{definition}
\label{material}
Let \cA\ be the conditional assertion \mbox{\ga \NI \gb}.
The material counterpart of \cA, denoted by $\tilde{\cal A}$, is the formula 
\mbox{$\alpha \ra \beta$}, where $\ra$, as usual, denotes material
implication.
If \bK\ is a set of assertions, its material counterpart $\tilde{\bf K}$
is the set of material counterparts of \bK.
\end{definition}

\begin{lemma}
\label{le:ex1}
Let \bK\ be a conditional knowledge base and \ga\ a formula.
Then $\tilde{\bf K} \models \alpha$ iff \bK\ preferentially entails
\mbox{{\bf true} \NI \ga}.
\end{lemma}

\proof
The {\em if} part follows from the fact that
any world satisfying $\tilde{\bf K}$ and not
\ga\ provides a one state preferential model satisfying \bK\ and not satisfying
\mbox{{\bf true} \NI $\alpha$}.
For the {\em only if} part suppose  $\tilde{\bf K} \models \alpha$.
By compactness, there is a finite subset of $\tilde{\bf K}$ that entails
$\alpha$. By rules {\bf S}, {\bf And} and 
{\bf Right Weakening} we conclude that \bK\ preferentially entails
\mbox{{\bf true} \NI \ga}.
\QED

\begin{corollary}
\label{co:exc}
Let \bK\ be a conditional knowledge base and \ga\ a formula.
The formula \ga\ is exceptional for \bK\
iff
$\tilde{\bf K} \models \neg \alpha$.
\end{corollary}
We see that, if \bK\ contains $n$ assertions, in the previous algorithm, 
we may go over the while loop
at most $O(n)$ times. Each time we shall have to consider at most
$n+1$ formulas and decide whether they are exceptional or not.
The whole algorithm needs at most $O(n^2)$ such decisions.
In the most general case all such decisions are instances of the
satisfiability problem for propositional calculus, therefore solvable
in non-deterministic polynomial time (in the size of the knowledge base \bK\ 
times the size of the formulas involved).
Therefore, even in the most general case, the problem is not much more complex
than the satisfiability problem for propositional calculus.
These results may be improved if we restrict ourselves to assertions
of a restricted type.
For example, if the assertions of \bK\ are of the Horn type
(we mean their material counterpart is a Horn formula), 
and the assertion \ab\ 
is of the same type, then, since each decision may be taken in polynomial
deterministic time, the whole algorithm runs in deterministic polynomial time.
The complexity discussion above is mainly of theoretical interest.
The important practical question is: given a fixed large knowledge base,
what information, of reasonable size, should be precomputed to allow
efficient answers to queries of the type: is an \cA\ in the rational closure?
The pre-computation of the different $C_n$ sub-bases would already
reduce the exponent of $n$ in the complexity of the algorithm by one.

J.~Dix noticed that the algorithm just presented for computing the rational 
closure of a finite knowledge base may be used to compute the preferential 
closure of such a knowledge base, since, by Corollary~\ref{co:reso} and 
Lemmas~\ref{co:acc} and~\ref{le:cperf}, the assertion \ab\ is in \bKp\ 
iff the assertion \mbox{$\alpha$ \NI {\bf false}} is in the rational
closure of the knowledge base \mbox{\bK\,$\cup \, \{ \alpha \NIm \neg \beta \}$}.

\subsection{A discussion of rational closure}
\label{subsec:disc}
We have so far shown that rational closure provides a mathematically
elegant and effective answer to the question of the title that satisfies
Thesis~\ref{rational}.
It is now time to evaluate whether it provides
an answer that matches our intuitions.
We shall first present two now classical knowledge bases, describe their 
rational closure and examine whether they fit our intuitions. 
Then, we shall discuss the way
rational closure treats inheritance of generic properties to abnormal 
individuals. Finally, we shall try to address the question of whether
our formalism is suitable to describe domain knowledge.
\begin{example}[Nixon diamond]
{\rm 
Let our knowledge base consist of the following two assertions.
\begin{enumerate}
\item {\em republican} \NI $\neg${\em pacifist}
\item {\em quaker} \NI {\em pacifist}
\end{enumerate} 
It is easy to see that none of the assertions of the base is exceptional,
but that the formula \mbox{{\em republican}$\wedge${\em quaker}} 
is exceptional.
From this we deduce that neither the assertion
\mbox{{\em republican}$\wedge${\em quaker} \NI {\em pacifist}} nor
the assertion
\mbox{{\em republican}$\wedge${\em quaker} \NI $\neg${\em pacifist}}
is in the rational closure.
This seems the intuitively correct decision in the presence of
contradictory information.
In fact, if we know somebody to be both a Quaker and a Republican,
we (i.e., rational closure) shall draw about him only conclusions that are 
logically implied by our information.
Rational closure endorses
\mbox{{\em worker} $\wedge$ {\em republican} \NI $\neg${\em pacifist}},
meaning that, since we have no information on the pacifism of workers,
we shall assume that Republican workers behave as Republicans in this
respect.
We (i.e., rational closure) also endorse 
\mbox{{\em pacifist} \NI $\neg${\em republican}}, 
meaning we are
ready to use contraposition in many circumstances.
We do not have \mbox{$\neg${\em pacifist} \NI {\em republican}}, though,
and quite rightly, since Republicans may well be a small minority
among non-pacifists.
We have 
\mbox{{\bf true} \NI $\neg$ ({\em republican} $\wedge$ {\em quaker})},
meaning we think being both a Republican and a Quaker is exceptional.
We endorse
\mbox{{\em republican} \NI $\neg${\em quaker}} and
\mbox{{\em quaker} \NI $\neg${\em republican}}, that are also intuitively
correct conclusions.
If we add to our knowledge base the fact that rich people
are typically Republicans, we shall deduce that rich people are typically
not pacifists, meaning we endorse a restricted form of transitivity.
We shall also deduce that Quakers are typically not rich, which is
perhaps more debatable.
We shall not conclude anything about the pacifism of rich Quakers though,
since rich Quakers are exceptional.
We shall not conclude anything either concerning rich Quakers that are 
not Republicans, which is more debatable.
If we want to conclude that rich non-Republican Quakers are pacifists,
we should add this assertion explicitly to the knowledge base.
The addition will not interfere with previously discussed assertions.
}
\end{example}
\begin{example}[Penguin triangle]
{\rm
Let our knowledge base consist of the following three assertions.
\begin{enumerate}
\item {\em penguin} \NI {\em bird}
\item {\em penguin} \NI $\neg${\em fly}
\item {\em bird} \NI {\em fly}
\end{enumerate}
The first two assertions are exceptional, the last one is not.
It follows that we (i.e., rational closure) endorse the following assertions:
\mbox{{\em fly} \NI $\neg${\em penguin}} (a case of contraposition),
\mbox{$\neg${\em fly} \NI $\neg${\em bird}} (another case of contraposition),
\mbox{$\neg${\em fly} \NI $\neg${\em penguin}} (penguins are exceptional,
even among non-flying objects),
\mbox{{\em bird} \NI $\neg${\em penguin}} (penguins are exceptional birds),
\mbox{$\neg${\em bird} \NI $\neg${\em penguin}} (penguins are exceptional
also among non-birds),
\mbox{{\em bird}$\wedge${\em penguin} \NI $\neg${\em fly}} 
(this is an intuitively correct preemption: we prefer specific information to 
non-specific information),
\mbox{{\em penguin}$\wedge${\em black} \NI $\neg${\em fly}} (black penguins
don't fly either, since they are normal penguins),
\mbox{{\em bird}$\wedge${\em green} \NI {\em fly}} (green birds are normal 
birds).

The following assertions are not endorsed:
\mbox{${\em bird} \wedge \neg {\em fly}$ \NI {\em penguin}} (there could be
non-flying birds other than penguins),
\mbox{${\em bird} \wedge \neg {\em fly}$ \NI $\neg${\em penguin}} (seems
intuitively clear),
\mbox{{\em penguin} \NI {\em fly}} (obviously).
}
\end{example}
A more general reflexion suggests the following.
Theorem~\ref{the:ratclos} shows that, in the rational closure,
no information about normal cases may be relevant to abnormal cases.
It is a very intriguing question whether human beings obey this rule
of reasoning or not. 
A specific example has been discussed by J.~Pearl in a personal communication.
It probably goes back to A.~Baker. 
Suppose we know that most Swedes are blond and tall. 
If we are going to meet Jon, whom we know to be short and to come
from Sweden, should we necessarily expect him to be fair? 
The answer endorsed by rational closure is {\em not necessarily},
since short Swedes are exceptional and we have no specific information about
such cases.
We do not know how people generally handle this and, even if we knew, 
it is not clear that AI systems should react in exactly the same way:
people are, after all, notoriously bad with statistical information.
The answer to the question how should people behave in this case, 
if they were smart and had all the relevant information, 
depends on the sociobiology of the Swedish population and is not relevant 
either.
There is very solid ground, though, to claim that, in the framework described
here, in which a knowledge base contains only positive conditional
assertions, the only sensible way to handle this problem is not to
expect anything about the color of Jon's hair. 
The reason is that, if we ever find out that most short Swedes are blond 
(or dark, for that matter)
it will be easy enough to add this information to our knowledge base.
On the contrary, had we chosen to infer that Jon is expected to be blond,
and had we found out that half of the short Swedes only are fair, we would
not have been able to correct our knowledge base to remove the unwanted 
inference:
adding the fact that most short Swedes are not blond being obviously 
incorrect.

Since, by looking at a number of examples, we have gathered some experience
on the behavior of rational closure, we would like to propose the following
strengthening of Thesis~\ref{rational}.
\begin{thesis}
\label{super}
The set of assertions entailed by any set of assertions \bK\ is a
rational superset of \ \oK.
\end{thesis}
Thesis~\ref{super} means that a reasonable system should endorse any
assertion contained in the rational closure, but it may also endorse
some additional assertions, as long as it defines a rational relation.
The search for natural constructions satisfying Thesis~\ref{super}, but
providing more {\em inheritance} than rational closure is open.

The main question that has not been addressed yet is whether conditional
knowledge bases are suitable to describe domain knowledge. 
Undoubtedly much work still has to be done before we may answer this question
satisfactorily. 
We shall only try to express here why we think the answer may well be 
positive.
Representing common sense knowledge is far from trivial in any one of the
existing formalisms, such as Circumscription or Default Logic.
Indeed to represent any substantive piece of common sense knowledge
in one of those formalisms, one needs to be an expert at the mechanics of
the formalism used, and they differ greatly from one formalism to the next.
Deciding on the different abnormality predicates in Circumscription
and the relations between them, or working out the default rules in Default 
Logic so as to ensure the correct precedence of defaults needs the hand 
of an expert.
In the formalism proposed here, conditional knowledge bases, the treatment
is much simpler since abnormality predicates do not appear explicitely
and the default information is described in a much poorer language than
Default Logic. 
We rely on the general algorithm for computing rational closure (or some other
algorithm that will be found suitable) to deal in a mechanical, uniform and
tractable manner with the interactions between different pieces of default
information. The fact that our language of assertions is much poorer than
other formalisms seems to us to be a great asset.

Nevertheless, it is probable that the size of useful conditional knowledge 
bases will be very large. Indeed, in our approach, adding new assertions
to the knowledge base may solve almost any problem.
Two main topics for further research may then be delineated.
The first one is to find practical ways to avoid having to look at the whole
knowledge base before answering any query. The set of assertions constituting
a knowledge base will have to be structured (off-line, once and for all) 
in such a way that irrelevant assertions do not have to be looked at.
The second one is to find lucid and compact descriptions of large conditional
knowledge bases. This will involve looking seriously into the question:
where does the conditional knowledge come from? Different answers may be 
appropriate in different domains: it may well be that conditional knowledge
is derived from causal knowledge in ways that are different from those
in which it is derived from conventions of speech or statistical information.

\section{Conclusion}
\label{sec:con}
We have presented a {\em mathematically tractable} framework for nonmonotonic
reasoning that can be proved to possess many pragmatically attractive 
features.
Its computational complexity compares favorably with that of most 
well-established systems.
In many cases the intuitively correct answer is obtained.
In others, the answer given and the way it was obtained provide an
interesting point of view on the knowledge base.
Much more practical experience is needed before one may assess the pragmatic
value of the approach.
The task of extending the results presented here to first-order languages
is not an easy one. First steps towards this goal are described 
in~\cite{LMTARK:90}.

\section{Acknowledgements}
\label{sec:Ack}
David Makinson suggested importing the thesis {\bf CV} of conditional logic
into the study of nonmonotonic consequence relations, i.e., suggested
to consider what is called here rational relations.
He conjectured that the corresponding family of models was that of 
ranked models.
He was also instrumental in stressing the importance of studying global
properties of nonmonotonic inference operations.
Discussions with the following people helped us to disprove hasty conjectures,
putting this work in perspective, and improve the presentation of this paper:
Johan van Benthem, Michael Freund, Haim Gaifman, Hector Geffner, 
Matthew Ginsberg, David Israel,
Sarit Kraus, John McCarthy and Judea Pearl.
Karl Schlechta's suggestions and J\"{u}rgen Dix's remarks on a previous draft
have been very useful. Finally, this paper has been fortunate
to receive attention and care of a rare quality from two anonymous referees.
We want to thank them.

\appendix
\section{Lemmas needed to prove Theorem~\protect\ref{comthe:rat}}
\label{appen:rep}
Let us suppose that some rational consequence relation \NI is given.
The notion of a consistent formula has been presented in 
Definition~\ref{def:cons}.
Let $S$ denote the set of all consistent
formulas.
Let us now recall Definition~10 of \cite{KLMAI:89}.

\begin{definition}
\label{def:normworld}
The world \mbox{$m \in {\cal U}$} is a normal world for $\alpha$ iff 
\mbox{$\forall \beta \in L$} such that \mbox{$\alpha \NI \beta$}, 
\mbox{$m \models \beta$}.
\end{definition}

The following is an easy corollary of Lemma~8 of~\cite{KLMAI:89}.

\begin{lemma}
\label{le:cons}
A formula is consistent iff there is a normal world for it.
\end{lemma}

We shall now define a pre-order relation on the set $S$.

\begin{definition}
\label{leq}
Where \mbox{$\alpha , \beta \in S$},
we shall say that $\alpha$ is not more exceptional than $\beta$ and write 
\mbox{$\alpha \R \beta$} iff 
\mbox{$\alpha \vee \beta$ \notNI $\neg\alpha$}.
\end{definition}

\begin{lemma}
\label{trans}
The relation $\R$ is transitive.
\end{lemma}
\proof
Straightforward from Lemma~\ref{rankder}. The fact that the relation $\R$ 
was restricted to the set $S$ is not used here and $\R$ would have been
transitive also on the whole 
language ${\cal L}$.
\QED
\begin{lemma}
\label{total}
Let \mbox{$\alpha , \beta \in S$}. Either 
\mbox{$\alpha \R \beta$} or
\mbox{$\beta \R \alpha$} 
(or both). In particular $\R$ is reflexive.
\end{lemma}
\proof
The proof proceeds by contradiction. Suppose we have
\mbox{$\alpha \notR \beta$}
and \mbox{$\beta \notR \alpha$}.
Then we have 
\mbox{$\alpha \vee \beta$ \NI $\neg\alpha$} and
\mbox{$\alpha \vee \beta$ \NI $\neg\beta$}.
By {\bf And} and {\bf Reflexivity} we have
\mbox{$\alpha \vee \beta$ \NI $\neg \alpha \wedge \neg \beta \wedge
(\alpha \vee \beta)$}, and therefore
\mbox{$\alpha \vee \beta$ \NI {\bf false}} and, by Rule~\ref{eq-1},
\mbox{$(\alpha \vee \beta) \wedge \beta$ \NI {\bf false}}. 
Therefore
\mbox{$\beta$ \NI {\bf false}}, 
contradicting \mbox{$\beta \in S$}. 
The fact that $\R$ was restricted to $S$ is crucial here.
\QED
The following will be useful in the sequel.
\begin{lemma}
\label{nm}
If
\mbox{$\alpha \R \beta$}, 
any normal world for
$\alpha$ that satisfies $\beta$ is normal for $\beta$.
\end{lemma}
\proof
Suppose
\mbox{$\alpha \R \beta$}, 
$m$ is normal for $\alpha$
and satisfies $\beta$. Let $\gamma$ be such that \mbox{$\beta$ \NI $\gamma$}.
We must show that \mbox{$m \models \gamma$}. Since $m$ is normal for $\alpha$
and satisfies $\beta$, it is enough to show that
\mbox{$\alpha$ \NI $\beta \ra \gamma$}.
But, \mbox{$\beta$ \NI $\gamma$} implies, by {\bf Left Logical Equivalence},
\mbox{$(\alpha \vee \beta) \wedge \beta$ \NI $\gamma$}.
By the rule {\bf S} of \cite{KLMAI:89}, one then obtains
\mbox{$\alpha \vee \beta$ \NI $\beta \ra \gamma$}.
But, by definition of $\R$, 
\mbox{$\alpha \vee \beta$ \notNI $\neg\alpha$} and, by {\bf Rational Monotonicity}
one deduces
\mbox{$(\alpha \vee \beta) \wedge \alpha$ \NI $\beta \ra \gamma$}.
\QED
\begin{definition}
\label{equiv}
Let \mbox{$\alpha , \beta \in S$}. We shall say that $\alpha$ is as exceptional
as $\beta$ and write \mbox{$\alpha \sim \beta$} iff \mbox{$\alpha \R \beta$}
and \mbox{$\beta \R \alpha$}. 
\end{definition}
Since $\R$ is reflexive and transitive, the relation $\sim$ is an equivalence
relation.
The equivalence class of a formula $\alpha$ will be denoted by 
$\overline{\alpha}$ and $E$ will denote the set of equivalence classes of
formulas of $S$ under $\sim$.
We shall write \mbox{$\overline\alpha \leq \overline\beta$} iff
\mbox{$\alpha \R \beta$} and we shall write
\mbox{$\overline\alpha < \overline\beta$} iff
\mbox{$\overline\alpha \leq \overline\beta$}  and 
\mbox{$\alpha \not \sim \beta$}. This notation should cause no confusion with
a similar notation used with a different meaning, 
in the context of preferential relations, in~\cite{KLMAI:89} and in 
Section~\ref{subsec:prefent}.
By Lemmas~\ref{trans} and \ref{total}, the relation $<$ is a strict total order
on the set $E$. 
\begin{lemma}
\label{<}
Let \mbox{$\alpha , \beta$} be consistent formulas.
If \mbox{$\overline\beta < \overline\alpha$} then
\mbox{$\beta$ \NI $\neg\alpha$}.
\end{lemma}
\proof
The assumption implies that \mbox{$\alpha \notR \beta$},
i.e., \mbox{$\alpha \vee \beta$ \NI $\neg\alpha$}.
Rule~(\ref{eqzero}) implies the conclusion.
\QED
\begin{lemma}
\label{betabar}
Let \mbox{$\alpha , \beta$} be consistent formulas.
If there is a normal world for $\alpha$ that satisfies $\beta$, then
\mbox{$\overline\beta \leq \overline\alpha$}.
\end{lemma}
\proof
If there is a normal world for $\alpha$ that satisfies $\beta$,
then we conclude by Lemma~\ref{<} that \mbox{$\alpha$ \notNI $\neg\beta$}.
\QED
Let $W$ be the ranked model \mbox{$\langle V , l , \prec \rangle$},
where $V \subseteq {\cal U} \times S$ is the set of all pairs
\mbox{$< m , \alpha >$} such that $m$ is a normal world for $\alpha$,
$l(<m,\alpha>)$ is defined to be $m$ and $\prec$ is defined as
\mbox{$<m , \alpha> \prec <n , \beta> $} iff 
\mbox{$\overline\alpha < \overline\beta$}.
To show that $W$ is a ranked model, we must prove that it satisfies the 
smoothness condition.
\begin{lemma}
\label{minchar}
In $W$, the state 
\mbox{$<m , \alpha>$} 
is minimal in 
$\widehat\beta$
iff 
\mbox{$m \models \beta$} 
and 
\mbox{$\overline\beta = \overline\alpha$}
\end{lemma}
\proof
First notice that 
\mbox{$<m , \alpha > \in \widehat\beta$} iff \mbox{$m \models \beta$}.
For the {\em only if} part, suppose that 
\mbox{$<m , \alpha>$} 
is minimal in 
$\widehat\beta$.
The world $m$ is normal for $\alpha$ and satisfies $\beta$.
By Lemma~\ref{betabar} we conclude that 
\mbox{$\overline\beta \leq \overline\alpha$}.
But, since $\beta$ is consistent, by Lemma~\ref{le:cons} there is a normal
world $n$ for $\beta$.
The pair \mbox{$<n , \beta>$} is an element of $V$ that satisfies $\beta$
and, by the minimality of \mbox{$<m , \alpha>$} in $\widehat\beta$,
\mbox{$<n , \beta> \not \prec <m , \alpha>$}, i.e., 
\mbox{$\overline\beta \not < \overline\alpha$}, i.e.,
\mbox{$\overline\alpha \leq \overline\beta$}. We conclude 
\mbox{$\overline\beta = \overline\alpha$}.
For the {\em if} part, suppose that 
$m$ is a normal world for $\alpha$ that satisfies $\beta$ and that
\mbox{$\overline\beta = \overline\alpha$}.
If $n$ is normal for $\gamma$ and 
\mbox{$<n , \gamma> \prec <m , \alpha>$} then 
\mbox{$\overline\gamma < \overline\alpha$} and therefore
\mbox{$\overline\gamma < \overline\beta$}.
By Lemma~\ref{<} \mbox{$\gamma$ \NI $\neg\beta$} and
$n$, which is normal for $\gamma$, cannot satisfy $\beta$.
The state 
\mbox{$<m , \alpha>$} is then minimal in $\widehat\beta$.
\QED
The following is an immediate corollary of Lemma~\ref{minchar}.
\begin{corollary}
\label{enough}
If $m$ is a normal world for $\alpha$ the pair 
\mbox{$<m , \alpha>$} 
is a state of $V$ and is minimal in $\widehat\alpha$.
\end{corollary}
\proof
Suppose $m$ is normal for $\alpha$.
First, since there is a normal world for $\alpha$, $\alpha \in S$ and
the pair \mbox{$<m , \alpha>$} is in $V$.
Since $m$ is normal for $\alpha$ it satisfies $\alpha$.
\QED
We may now prove that the model $W$ satisfies the smoothness property and
defines the consequence relation \NI.
\begin{lemma}
\label{smoothness}
Let $\alpha$ be a consistent formula. The set 
\mbox{$\widehat\alpha \subseteq V$} is smooth.
\end{lemma}
\proof
Suppose \mbox{$<m , \beta> \in \widehat\alpha$}. Then, $m$ is a normal world
for $\beta$ that satisfies $\alpha$ and,
by Lemma~\ref{betabar}, \mbox{$\overline\alpha \leq \overline\beta$}.
If 
\mbox{$\overline\alpha = \overline\beta$}, 
then, by Lemma~\ref{minchar},
\mbox{$<m , \beta>$} is minimal in $\widehat\alpha$.
Otherwise, \mbox{$\overline\alpha < \overline\beta$}.
In this case, let $n$ be any world normal for $\alpha$ (there is such a 
world since $\alpha$ is consistent).
The pair \mbox{$<n , \alpha>$} is minimal in $\widehat\alpha$ by 
Lemma~\ref{minchar} and 
\mbox{$<n , \alpha> \prec <m , \beta>$}.
\QED
\begin{lemma}
\label{nottoomany}
If 
\mbox{$<m , \alpha>$} 
is minimal in $\widehat\beta$, then $m$ is normal 
for $\beta$.
\end{lemma}
\proof
Suppose 
\mbox{$<m , \alpha>$} is minimal in $\widehat\beta$.
By Lemma~\ref{minchar}
\mbox{$\overline\alpha = \overline\beta$}.
Therefore
\mbox{$\alpha \R \beta$}.
But $m$ is normal for $\alpha$ and satisfies $\beta$, and Lemma~\ref{nm}
implies that $m$ is normal for $\beta$.
\QED

\section{Non-standard probabilistic semantics}
\label{appen:nonstandard}
\subsection{Introduction}
We shall describe now, in Definition~\ref{def:ourmodels} 
another family of probabilistic models,
they provide much more direct semantics for nonmonotonic reasoning than
Adams', at the price of using the language of non-standard 
(in the sense of A.~Robinson) probability theory.
The purpose of this section is to provide additional evidence in support 
of Thesis~\ref{rational}.
We shall show that rational relations are exactly those that may be defined 
by non-standard {\em probabilistic} models.
In other terms, if, given a probability distribution, we decide to
accept the assertion \ab\ iff the conditional probability of \gb\ given \ga\
is very close to one, then the consequence relation we define is rational.
On the other hand, any rational relation may be defined, in such a way,
by some probability distribution.
The results presented in the appendix are not used in the body of the paper.
A different representation theorem for rational relations, also based
on Theorem~\ref{comthe:rat}, in terms of one-parameter families of
standard probabilistic models has been proved recently by 
K.~Satoh~\cite{Satoh:89}. Results relating the semantics of conditionals
and non-Archimedean probabilities seem to have been obtained by R.~Giles
around 1980.

There is a school of thought in Artificial Intelligence,
represented in particular by~\cite{Che:88,Che:88b},
that denies the validity of the logical approach to modeling common-sense 
reasoning.
The alternative suggested is the Bayesian probabilistic approach.
Namely, the only way in which we should make sensible inferences
from our knowledge \ga\ is by estimating the conditional probability of the
required conclusion \gb\ given our knowledge \ga, 
and then adopting \gb\ if we are satisfied that this conditional probability 
is close enough to $1$.
We believe that this approach may run into considerable 
practical difficulties, the choice being between keeping an explicit data 
base of these many conditional probabilities or
estimating them from a small sample.
The chief source of difficulty here is that knowing the probability of 
\ga\ and \gb\ tells you very little about the probability of their 
intersection.

But we shall not argue the matter in detail here.
The main purpose of this section is to show that rational knowledge bases 
may be considered to come from such a probabilistic model, 
if we let the cut-off point of how close the conditional probability of 
\gb\ given \ga\ has to be before we are ready to adopt \gb\ as a sensible 
consequence of \ga, 
approach $1$ as a limit.
Namely, \gb\ is a sensible consequence of \ga,
iff the conditional probability is {\em infinitesimally} close to $1$.
In order to have an interesting theory, there must be probabilities 
that are not standard real numbers,
but belong to a richer system of numbers, containing some infinitesimally 
small numbers.

We shall show that this approach allows one to keep a probabilistic 
intuition while thinking about common-sense reasoning,
namely think about $\alpha\NI\beta$ as meaning
that the conditional probability of \gb\ given \ga\ is large,
and still defines a well-behaved consequence relation that is not necessarily 
monotonic.
Note that if one considers a standard probabilistic model
and accepts $\alpha\NI\beta$ as satisfied by the model 
iff the conditional probability $Pr(\alpha/\beta)>1-\epsilon$, 
for some choice of a positive $\epsilon$
one obtains a consequence relation that is not well-behaved.
For instance, one may have $\alpha\NI\beta$ and $\alpha\NI\gamma$ 
satisfied by the model,
while $\alpha \NI\ \beta \wedge \gamma$ is not satisfied.
If, on the other one hand, one chooses $\epsilon$ to be $0$,
one obtains a well-behaved consequence relation, but this relation 
is always monotonic, and the entailment defined is classical entailment
(read \NI\ as material implication).
J.~McCarthy told us he suggested considering non-standard probabilistic
models long ago, but, as far as we know, this suggestion has not been
systematically pursued.

The structure of this section is as follows:
first we shall,
briefly, survey the basic notions of non-standard analysis.
We shall also introduce non-standard probability spaces.
Then we shall introduce non-standard probabilistic models 
for non-monotonic reasoning,
define the consequence relation given by such a model 
and prove that any consequence relation given by a non-standard probability 
model is rational.
Lastly we shall show that the axioms are complete for this interpretation,
i.e., any rational consequence relation can be represented as the 
consequence relation given by some non-standard probability model,
at least in the case the language ${\cal L}$ is countable. 
If ${\cal L}$ is not countable, an easy counter-example shows the result does 
not hold, but we shall not elaborate in this paper.

\subsection{Non-Standard Analysis}
Non-standard analysis was invented by Abraham Robinson 
in order to give a rigorous development of analysis in which 
limiting processes are replaced by behaviour at the infinitesimally small, 
e.g., the derivative becomes a quotient of the change in
the function divided by the change in the argument, 
when the argument is infinitesimally increased.
In this section we shall give a very brief introduction to the basic ideas.
The reader interested in a full treatment can consult 
A.~Robinson's~\cite{Robinson:66} or Keisler's~\cite{Keisler:76} books
on the topic.
More advanced topics related to non-standard probability theory are surveyed 
in~\cite{Cutland:83}.

The basic idea of non-standard analysis is to extend the real numbers to  a
larger ordered field while preserving many of the basic properties 
of the reals.
Therefore, we consider a structure of the form
\[ {\cal R}\sstar=\langle R\sstar,+\sstar,\times\sstar,<\sstar,0,1\rangle \]
such that ${\cal R}\sstar$ is an elementary extension of the standard real
numbers, namely ${\bf R} \subset R\sstar$, 
the operations and the order relation of \calR\
extend those of {\bf R} and for every first order formula $\Phi$
\[ \calR\models\Phi(x_{1},\ldots,x_{n})\ \mbox{iff}\
 {\bf R} \models\Phi(x_{1},\ldots,x_{n}) \]
for $x_{1},\ldots,x_{n}\in {\bf R}$.
Since we would like to consider not only properties of the real numbers,
but real valued functions, functions from real valued functions into reals, 
and so on, we shall consider a richer structure: 
the superstructure of the real numbers.
\begin{definition} 
The superstructure of the set $X$ is
 $V_{\infty}(X)=\bigcup_{n=0}^{\infty}V_{n}$
where $V_{n}$ are defined by induction:
\begin{itemize}
    \item $V_{0}=X$
    \item $V_{n+1}={\cal P}(V_{n})\cup V_{n}$ where ${\cal P}(Y) $ 
is the power set of Y.
\end{itemize}
\end{definition}
Note that the superstructure of $X$ contains all the relations on $X$, 
all $n$-valued functions from $X$ into $X$, etc.
In a non-standard model of the real numbers we would like to have a 
non-standard counterpart to any standard member of the superstructure 
of the real numbers.
Note that the set theoretical relation $\in$ makes sense 
in the superstructure of $X$.
Recall that a formula
of the first order language having only $\in$ as a non logical constant is
called bounded if is constructed by the usual connectives and {\em bounded }\
quantifiers, 
namely $(\forall x \in y)$ and $(\exists x \in y)$ meaning respectively: 
$\forall x \ \mbox{\em if} \ x \in y \ \mbox{\em then}
\ldots$ and $\exists x \ x \in y \wedge\ldots$.
\begin{definition}
A non-standard model of analysis is an ordered field \calR 
that is a {\em proper} extension of the ordered field of the reals, 
together with a map $\sstar$ from the superstructure of {\bf R}
into the superstructure of {\calR}, such that for every {\em bounded} formula
$\Phi(x_{1} \ldots x_{n})$:
\[V_{\infty}({\bf R})\models\Phi(a_{1}\ldots a_{n})\ \mbox{iff}\
 V_{\infty}(\calR)\models\Phi(a_{1}\sstar
\ldots a_{n}\sstar) \ {\bf (Leibniz \ Principle)}\]
and such that for $x \in {\bf R}$  $x\sstar = x$
(we assume that $\sstar$ transforms the standard
operations of {\bf R} into those of \calR).
\end{definition}
The Leibniz principle guarantees that the non-standard counterpart of any
standard notion (namely its $\sstar $) preserves many of the properties 
of the standard object.
In particular it is an object of the same kind:
for example if A is a set of functions
from {\bf R} to {\bf R}, then $A\sstar$ is a set of functions 
from {\calR} into {\calR}.
As another example consider the absolute value as a function 
from {\bf R} to {\bf R}.
In {\bf R} it has the property
\[ \left ( \forall x \in {\bf R}) (\absv{x} \geq 0 \wedge \left ( \absv{x} = 0 
\leftrightarrow x = 0 \right ) \right ) \]
Then by the Leibniz principle 
\[(\forall x \in \calR) 
\left ( \absv{x}\sstar \geq\sstar0 \wedge \left ( \absv{x}\sstar = 0 
\leftrightarrow x = 0 \right ) \right ) \]
In fact since the $\sstar$ versions of the standard arithmetic operations and
relations (like $\leq$, $\geq, >, <$) are so similar to the standard ones 
(they extend them) we shall simplify the notation by dropping the $\sstar$, 
letting the context determine whether we mean the standard operation on 
{\bf R}, or its extension to {\calR}.
The next theorem shows that this is not a formal game:
\begin{theorem}[Robinson]
 There exists a non-standard model for analysis.
\end{theorem}
 The proof is an application of the compactness theorem.
The extension of {\bf R}, {\calR}, is not unique  but nothing in the following
arguments depends on the particular choice of the non-standard extension 
of {\bf R}.
So fix one such extension {\calR}.
\begin{definition}
\begin{enumerate}
 \item $x \in \calR$,  $x \neq 0$, is called {\em finite} if 
$\absv {x} < y$ for some $y \in {\bf R}$, or, equivalently, 
if $\absv {x} < n$ for some natural number $n$.
 \item $x \in \calR$ is called {\em infinitesimal} 
if for all $\epsilon$ in {\bf R}, $\epsilon>0$, $\absv{x}<\epsilon$. 
Following our definition $0$ is infinitesimal.
 \item $x\in V_{\infty}(\calR)$ is called {\em internal} if
 $x\in y\sstar$ for some $y\in V_{\infty}({\bf R})$.
The set of internal objects is denoted by $V_{\infty}\sstar$.
\item $x\in V_{\infty}(\calR)$ is standard if $x=y\sstar$ for some $y\in
 V_{\infty}({\bf R})$.
\end{enumerate}
  \end{definition}
It follows easily, from the fact that {\calR} is a proper 
extension of {\bf R}, 
that there
are infinitesimal, as well as infinite, members of {\calR}.
In fact $x$ is infinitesimal iff $1/x$ is infinite.
If \/{\bf N} is the set of natural numbers, one can show that \/{\bf N} is a
 proper
subset of \/${\bf N}\sstar$ and every member of \/${\bf N}\sstar-{\bf N}$ 
is called a non-standard natural number.
\begin{lemma}\label{lemma:sum}\begin{enumerate}
 \item The sum, product and difference of two infinitesimals is infinitesimal.
 \item The product of an infinitesimal and a finite member of \/{\calR} is
infinitesimal.
\end{enumerate}\end{lemma}
\begin{theorem}[Robinson's Overspill Principle]\label{Robinson-overspill}
Let $\langle A_{n}\mid n\in {\bf N} \rangle$ be a sequence of members of 
$V_{k}({\bf R})$ for some $k\in {\bf N}$.
Assume also that, for all $n \in {\bf N}$,  $A_{n} \neq \emptyset$ and
$A_{n+1}\subseteq A_{n}$. 
Then $\bigcap_{n\in {\bf N}}A_{n}\sstar$ is not empty.
\end{theorem}
{\bf Sketch of proof :} 
Note that a sequence of elements of $V_{k}({\bf R})$ can be considered
to be a function from {\bf N} into $V_{k}({\bf R})$, 
and therefore it is a member of $V_{\infty}({\bf R})$. 
Hence $\langle A_{n}\mid n\in {\bf N} \rangle\sstar$ makes sense and it is a
function from ${\bf N}\sstar$ into $V_{k}({\bf R})\sstar$.
Its value at $h\in {\bf N}\sstar$ will be denoted by $(A)\sstar_{h}$.
Note that $(A)\sstar_{n}=A_{n}\sstar $ for $n\in {\bf N}$.
Let $h\in {\bf N}\sstar-{\bf N}$.
One can easily check, using the Leibniz principle,
that $(A)\sstar_{h}$ is not empty and that for 
$n\in {\bf N}$ $(A)\sstar_{h}\subseteq A\sstar_{n}$, 
hence $\bigcap_{n\in {\bf N}}A_{n}\sstar$ is not empty.
\QED
We can now define the notion of non-standard probability space, which is like
a standard (finitely additive) probability space, 
except that the values of the probability function are in {\calR}.
\begin{definition} An {\calR}-{\em probability space} is a triple
 $\langle X, {\cal F}, Pr\rangle$ where X is a non-empty set , $\cal F$ is a
 Boolean
subalgebra of ${\cal P}(X)$, (namely $X\in\cal F$, 
$\emptyset\in\cal F$, and $\cal F$
is closed under finite unions, intersections and differences)
 and $Pr$ is a function from ${\cal F}$ into {\calR} such that
\begin{enumerate}
 \item $Pr(A)\geq0$ for $A\in\cal F$.
\item $Pr(X)=1$
\item $Pr(A\cup B)=Pr(A)+Pr(B)$ for $A, B\in \cal F$, A and B disjoint
\end{enumerate}
\end{definition}
Note that many of the notions that are usually associated 
with probability spaces are immediately generalized to 
{\calR}-probability space, 
like independence of `events' (namely sets in $\cal F$) 
and conditional probability: 
if $Pr(A)\not=0$
then the conditional probability of B given A, is 
\[Pr(B \mid A)=\frac{Pr(A \cap B)}{Pr(A)}.\]
See \cite{Cutland:83} for sophisticated applications of 
non-standard probability spaces.
A useful way of getting {\calR}-probability spaces is by using 
hyperfinite sets, 
sets which are considered by {\calR} to be finite.
\begin{definition} 
An internal object $A\in V\sstar_{\infty} $ is called {\em hyperfinite} 
iff there
exists a function $f\in\vstar$ and $h\in {\bf N}\sstar$ 
such that f is a 1-1 mapping of h onto A.
Note that we follow the usual set theoretical convention by which
a natural number is identified with all smaller natural numbers.
Of course here we apply this convention also to non-standard natural numbers.
\end{definition}
By applying the Leibniz principle we can show that if A is hyperfinite and 
B is an internal subset of A, then B is hyperfinite.
Given an {\calR}-valued function f which is
internal, and A an hyperfinite subset of the domain of f, we can naturally
define the `sum' of the values of f on A, 
$\sum\sstar_{x\in A}f(x)$.
\sumstar\ is defined by taking the $\sstar$ of the standard operation 
of taking the sum of a finite set of real numbers.
\sumstar\ shares many of the properties of its standard counterpart, 
for example
\[\sumstar_{x\in A\cup B}f(x)=\sumstar_{x\in A}f(x)+\sumstar_{x\in B}f(x)\]
for A, B hyperfinite and disjoint.
The next definition generalizes the notion of a finite probability space.
\begin{definition}[Hyperfinite Probability Space]  
\label{def:HFPS}
Let $A\in\vstar$ be an hyperfinite set, 
let f be an internal {\calR}-valued function on A, 
which is not constantly zero and such that for $x\in A$ $f(x)\geq0$.
Then  the {\em {\calR}-probability
space generated by A and f} (denoted by $PR\sstar(A, f)$ ) is
$ \langle A, {\cal F}, Pr\rangle$
where $\cal F$ is the collection of all internal subsets of A, 
and ${\em Pr}$ is given by
\[ Pr(B)=\frac{\sumstar_{x\in B}f(x)}{\sumstar_{x\in A}f(x)} \]
\end{definition}
One can verify that under the conditions of Definition~\ref{def:HFPS}, 
$PR\sstar(A, f)$ is a {\calR}-probability space.

\subsection{Non-standard Probabilistic Models and Their Consequence Relations}
\label{sec:NSPM}
An {\calR} probabilistic model is an {\calR}-probability measure 
on some subset \cM\ of \cU.
Of course, we assume that for every formula of our language, \ga, the set
$\hat{\ga}$ is measurable, namely it is in $\cal F$.
The probability measure induces a non-standard probability assignment
to the formulas of the language by $Pr(\ga)=Pr(\hat{\ga})$.
The {\calR} probabilistic model {\cM} is said to be {\em neat} 
if for every formula, \ga, if $Pr(\ga)=0$ then
{\ga} is satisfied in no world of {\cM}.
\begin{definition}
\label{def:ourmodels}
\begin{enumerate}
\item  Let {\cM} be an {\calR} probabilistic model.
The conditional assertion
$\ga\NI\gb$ is valid in {\cM}, $\cM\models\ab$,  
if either $Pr(\ga)=0$ or the conditional probability
of {\gb} given {\ga} is infinitesimally close to 1, i.e.,
$1-Pr(\gb \mid \ga)$ is infinitesimal.
Note that this is equivalent to saying that $Pr(\alpha) = 0$ or
{\mbox{$Pr(\neg\gb\mid\ga)$}} is infinitesimal.
\item The consequence relation defined by {\cM} is:
\[ K(\cM)=\{\ab\mid\cM\models\ab\} \]
\end{enumerate}
\end{definition}
\begin{theorem}[Soundness for Non-standard Probabilistic Models]
\label{SoundnessProb}
For every {\calR} probabilistic model {\cM}, K(\cM) is a rational consequence
relation.
\end{theorem}
\proof 
{\bf Left Logical Equivalence}, {\bf Right Weakening}, and 
{\bf Reflexivity} are immediate.
{\bf And} follows from:
\[Pr(\neg (\gb\wedge\gc) \mid \ga) = Pr((\neg\gb\vee\neg\gc)\mid\ga)
\leq Pr(\neg\gb\mid\ga) + Pr(\neg\gc\mid\ga)\]
and from the fact that the sum of two infinitesimals is infinitesimal.
{\bf Or} is proved by the following manipulation:
\begin{eqnarray}
Pr(\neg\gc\mid\ga\vee\gb)&=&\frac{Pr(\neg\gc\wedge(\ga\vee\gb))}{Pr(\ga\vee\gb)}
\leq \nonumber \\
\frac{Pr(\neg\gc\wedge\ga)}{Pr(\ga\vee\gb)} &+& \frac{Pr(\neg\gc\wedge\gb)}
{Pr(\ga\vee\gb)}\leq \nonumber \\
\frac{Pr(\neg\gc\wedge\ga)}{Pr(\ga)} &+& \frac{Pr(\neg\gc\wedge\gb)}
{Pr(\gb)} = Pr(\neg\gc\mid\ga) + Pr(\neg\gc\mid\gb)\nonumber
\end{eqnarray} 
and again using the fact that the sum of two infinitesimals is infinitesimal.
We assumed above that $Pr(\ga)>0$ and $Pr(\gb)>0$.
If this fails then the argument is easier.
We shall prove {\bf Rational Monotonicity} by contradiction, so we assume that
$\ga\NI\neg\gb$ is not in K(\cM), and that $\ga\NI\gc$ is in K(\cM).
We shall prove that $\ga\wedge\gb\NI\gc$ is in K(\cM).
We can assume that
 $Pr(\ga\wedge\gb)>0$ (hence $Pr(\ga)>0$) otherwise the argument is trivial.
\begin{eqnarray}\label{eq:RM}
Pr(\neg\gc\mid\ga\wedge\gb)&=&
\frac{Pr(\neg\gc\wedge\ga\wedge\gb)}{Pr(\ga\wedge\gb)}= \nonumber \\
 \frac{Pr(\neg\gc\wedge\ga\wedge\gb)}{Pr(\ga)}&/&
\frac{Pr(\ga\wedge\gb)}{Pr(\ga)}\leq\nonumber \\
\frac{Pr(\neg\gc\wedge\ga)}{Pr(\ga)}
&/&\frac{Pr(\ga\wedge\gb)}{Pr(\ga)}=\nonumber
\\
Pr(\neg\gc\mid\ga)&\times&\frac{1}{Pr(\gb\mid\ga)}
\end{eqnarray}
Since $\ga\NI\neg\gb$ is not in $K(\cM)$ , we get that $Pr(\gb\mid\ga)$ is not
infinitesimal, hence $\frac{1}{Pr(\gb\mid\ga)}$ is finite.
By Lemma~\ref{lemma:sum}
$Pr(\neg\gc\wedge\ga)\times\frac{1}{Pr(\gb\mid\ga)}$
is infinitesimal.
Hence by Equation~\ref{eq:RM}, $\ga\wedge\gb\NI\gc$ is in $K(\cM)$.
{\bf Cautious Monotonicity } now follows easily.
Suppose \mbox{$\alpha \NIm \gamma$} and \ab\ are both in $K(\cM)$.
If \mbox{$\alpha \NIm \neg \beta$} is not in $K(\cM)$, we conclude
by {\bf Rational Monotonicity}.
If \mbox{$\alpha \NIm \neg \beta$} is in $K(\cM)$, we must have 
$Pr(\ga) = 0$, since \mbox{$Pr ( \beta \mid \alpha )$} and
\mbox{$Pr ( \neg \beta \mid \alpha)$} cannot be both infinitesimally close
to 1.
Therefore \mbox{$Pr(\alpha \wedge \beta ) = 0$} and we conclude that
\mbox{$\alpha \wedge \beta \NIm \gamma \in K (\cM)$}.
\QED

\subsection{Completeness for the Non-Standard Probabilistic Interpretation}
\label{subsec:ComNSI}
\begin{theorem}\label{completeness-nonstd}
Suppose the language ${\cal L}$ is countable (this assumption cannot be 
dispensed with)
and $P$ is a rational consequence relation on ${\cal L}$.
Let {\calR} be any non-standard model of analysis, 
then there exists an {\calR}-probabilistic neat model {\cM} 
such that $K(\cM)=K$.
\end{theorem}

\proof 
Let $W=\langle S, l, \prec \rangle $, with ranking function $r$, be
a countable (i.e., $S$ is countable) ranked model that defines 
the consequence relation $P$. 
The model built in the proof of
Theorem~\ref{comthe:rat} shows that such models exist. 
If $S$ is finite, or even if each level in $W$ is finite and $W$ is well-founded, 
one may simply
use the construction described just before Lemma~\ref{le:unnamed}, 
with some arbitrary infinitesimal $\epsilon$.
In case the model $W$ is infinitely {\em broad}, i.e., has some level
containing an infinite number of states, then the construction has 
to be slightly more sophisticated, but the real difficulty appears when
$W$ is not well-founded, and we have already remarked that there are rational
relations that have no well-founded ranked model.
Following the proof of Lemma~\ref{le:unnamed} we would like to assign
a (non-standard) probability distribution to the states of the model in such a
way that the relative probabilty of a level to that of a lower level 
is infinitesimal,
but, for every formula which is satified at a given level, 
we would like to keep its relative weight within the level non infinitesimal.
To each formula we shall assign a positive real number $r$ such that, 
if the formula is satisfied at level $l$,
its relative probability within this level should be at least $r$.
In order that these requirements not be contradictory, 
the sum of the $r$'s so assigned should be at most $1$.
Quite arbitrarily, we pick for the $i$-th formula $r=1/2^{i+1}$.
Now we have to show that we can find a probability
assignment satisfying these requirements.
We shall define a set $B_{n}$ of all probability assignments 
that are {\em good up to rank $n$}.
An assignment which is {\em good for every $n$} 
will satisfy our requirements.
So, we would like to intersect the $B_{n}$'s.
The overspill principle will tell us that this intersection is not empty.

Since $S$ is countable
we may assume that \mbox{$S = {\bf N}$}.
Since every countable linear ordering may be order embedded into the real numbers, 
we may assume without loss of generality that the ranking
function, $r$, is into {\bf R}.
Since $\prec$ is a partial ordering of {\bf N}, $\prec\sstar$
is a partial ordering of ${\bf N}\sstar$ 
which is ranked by the ranking function
$r\sstar$ mapping ${\bf N}\sstar$ into {\calR}.

For each formula {\ga}, let \mbox{$A_{\ga}=(\hat{\ga})\sstar$}.
Note that $A_{\ga}$ is a subset of ${\bf N}\sstar$ (but must not be a subset
of {\bf N}).
We can now associate a world, $\cU_{h}$, with each 
\mbox{$h \in {\bf N}\sstar$}, 
defined by \mbox{$\cU_{h} \models p$} iff  \mbox{$h \in A_{p}$}.
It is easily checked that, for standard $h$ 
(i.e., $h\in {\bf N}$), one has \mbox{$\cU_{h} = l(h)$}
and that, for arbitrary $h$, 
\mbox{$\cU_{h} \models \ga$} iff \mbox{$ h \in A_{\ga}$}. 
Our idea now is to find an $h$ in ${\bf N}\sstar$ and an internal function
$f$, from $h$ into {\calR} such that, 
if we consider the probability distribution given
by the hyperfinite probability space \mbox{$PR\sstar(h, f)$} on the set of worlds
\mbox{$\{ \cU_{k} \mid k \in {\bf N}\sstar , k < h \}$}, 
we shall get a probabilistic model whose consequence relation is exactly $P$
(recall that we are identifying a member of ${\bf N}\sstar$ with 
the set of smaller members of ${\bf N}\sstar$).
Fix an enumeration
\mbox{$\langle \ga_{n}\mid n\in {\bf N}\rangle $} 
of all the formulas of our language.
For \mbox{$i\in {\bf N}$} let $x_{i}$ be the real
number such that the ranking function $f$ maps all the
states minimal in $\hat{\ga_{i}}$ to it.

We are now going to define a sequence of sets of possible approximations to
the object we are looking for, namely the appropriate $h\in {\bf N}\sstar$ and 
the appropriate $f$.
For \mbox{$n \in {\bf N}$}, let $B_{n}$ be the set of all triples of the form
$(k, \epsilon, f)$ that have the following properties:
\begin{enumerate}
\item $k\geq n$,
\item $\epsilon \in {\bf R}$, $\epsilon>0$, $\epsilon \leq1 / n$,
\item  $f$ is a function from {\bf N}  into {\bf R} such that for any
$s\in {\bf N}$, $f(s)>0$,
\item\label{itemsmall}
for any $x, y \in {\bf R}$ such that $x < y$, if $x$ and $y$ are in the range 
of the ranking function $r$ on $k$, then
\[ \frac{\sum_{m<k, r(m)=y}f(m)}{\sum_{m<k, r(m)=x}f(m)} \leq \epsilon , \]
\item  \label{itemstd} 
for $\ga_{i}$, \mbox{$i < k$}, if 
\mbox{$C = \hat{\ga_{i}} \cap \{ j \mid r(j) = x_{i} \}
\cap \{0, \ldots, k-1\} \neq \emptyset$},
then
\[\frac{\sum_{m \in C , m < k} f(m)}{\sum_{m < k , r(m) = x_{i}} f(m)}
\geq \frac{1}{2^{i+1}} \]
\end{enumerate}

It easily follows from the definition of the sequence of sets 
\mbox{$\langle B_{n} \mid n \in {\bf N} \rangle$} that 
\mbox{$B_{n+1} \subseteq B_{n}$} for \mbox{$n \in {\bf N}$}.
One may also verify from item~\ref{itemsmall} that, if 
\mbox{$(k, \epsilon, f) \in B_{n}$}
and if $x$ is in the range of $r$ on $k$, then:
\begin{equation}\label{eq:ratio-inf}
\frac{\sum_{m < k , r(m) > x} f(m)} {\sum_{m < k , r(m) = x} f(m)} \leq 
\sum_{i=1}^{\infty} \epsilon^{i} = \frac{\epsilon}{1-\epsilon}
\end{equation}
\begin{lemma}
\label{nonempty}
For any \mbox{$n \in {\bf N}$}, \mbox{$B_{n} \neq \emptyset$}.
\end{lemma}
\proof 
The proof is essentially similar to the remarks preceding the
proof of Lemma~\ref{le:unnamed} in Section~\ref{subsec:adams}.
Let, indeed, $W_{n}$ be the finite ranked model defined by
\mbox{$\langle \{0, \ldots , n-1\}, \prec, l\rangle$}.
We can easily arrange a probability assignment for it such that the ratio
of the probability of each rank and and each smaller rank will be at most
$1/n$.
Within the rank we have to satisfy item~\ref{itemstd} in the definition of
$B_{n}$ but we can easily arrange for $i<n$, that if $\ga_{i}$
has a non empty intersection with this rank, 
then its relative probability within this
rank is at least $\frac{1}{2^{i+1}}$.
This may be arranged because
$\sum_{i\in {\bf N}}\frac{1}{2^{i+1}}=1$.
If we extend this probability assignment to
any function from {\bf N} into {\bf R}, we see that 
\mbox{$(n , \frac{1}{n}, f) \in B_{n}$}.
\QED
Once we have Lemma~\ref{nonempty} we can use Robinson's overspill principle
(Theorem ~\ref{Robinson-overspill}) to show that
$\cap_{n\in {\bf N}}B_{n}\sstar$ is not empty.
So let \mbox{$(\tilh, \tilep, \tilf)$} be a member of $B_{n}\sstar$ 
for every $n\in {\bf N}$.
One can easily verify that {\tilh} is in ${\bf N}\sstar$ and that it is 
a non-standard natural number: 
indeed for every \mbox{$n \in {\bf N}$},  \mbox{$\tilh > n$} since
\mbox{$(\tilh, \tilep, \tilf)\in B_{n}\sstar$}.
Similarly {\tilep} is a positive member of {\calR} such that for every
standard natural number $n$ we have $\tilep\leq\frac{1}{n}$, 
hence {\tilep} is a positive infinitesimal.
Also {\tilf} is a function from ${\bf N}\sstar$ into the positive
members of {\calR}, 
satisfying the appropriate transfer of items~\ref{itemstd} and
\ref{itemsmall} into the context of {\calR}.
In particular, Equation~\ref{eq:ratio-inf} carries over and we have
\mbox{$x = r\sstar(m)$} for some \mbox{$m \in {\bf N}\sstar , m < \tilh$}:
\begin{equation}\label{eq:ratio2}
\frac{\sumstar_{m < \tilh , r\sstar(m) > x} \tilf(m)}
{\sumstar_{m < \tilh , r\sstar(m) = x} \tilf(m)} \leq\frac{\tilep}{1-\tilep}
\end{equation}
We conclude therefore that the left-hand side of Equation~\ref{eq:ratio2} 
is infinitesimal.
We claim that the {\calR} probabilistic model {\cM} whose collection of states
is {\tilh}, i.e.,
\mbox{$\{ m \mid m \in {\bf N}\sstar, m < \tilh \}$}, 
the world associated with $m$ is $\cU_m$, and the probability measure is given
by the hyperfinite probability space $PR\sstar(\tilh, \tilf)$ is the model
we are looking for.
Since any $f$ satisfying the requirements can be multiplied by 
any positive member of {\calR } and still satisfies the requirements, 
we may assume without loss of generality that
\[\sumstar_{m\in {\bf N}\sstar , m < \tilh} \tilf(m)=1 \]
Note that {\cM} is a neat model since, if we have both \mbox{$m < \tilh$} and 
\mbox{$\cU_{m} \models \ga$},
we must also have \mbox{$Pr(\ga) \geq \tilf(m) > 0$}.
\begin{claim}
\label{main-claim}
$K(\cM)=K$
\end{claim}
\proof
First note that \mbox{$\gafalse \in K(\cM)$} iff \mbox{$\gafalse \in K$}.
If \mbox{$\gafalse \in K$}, then \mbox{$A_{\ga} = \emptyset$}, hence 
\mbox{$\{m \mid m \in {\bf N}\sstar, \cU \models \ga \} = \emptyset$}.
Therefore \mbox{$\gafalse \in K(\cM)$}.
For the other direction, if \mbox{$\gafalse \not\in K$}, 
then, for some \mbox{$m \in {\bf N}$},
\mbox{$U_{m} \models\ga$}. 
But $m$, being a standard natural number, is less than {\tilh}, 
hence some state in {\cM} satisfies {\ga}.
By the neatness of {\cM}, \mbox{$\gafalse \not\in K(\cM)$}.
By the previous remark, we can now assume that 
\mbox{$\gafalse \not\in K$}, hence \mbox{${\gahat \neq \emptyset}$}.
Let \mbox{$m \in {\bf N}$} be minimal in {\gahat} and let \mbox{$x = r(m)$}.
Let \mbox{$i\in {\bf N}$} be such that \mbox{$\ga = \ga_{i-1}$}.
In particular we have:
\[ (\forall y \in {\bf R}) (y < x \Rightarrow r^{-1}(y) \cap \gahat) = 
\emptyset . \]
Using the Leibniz principle we get:
\[ \{ m \mid m \in {\bf N}\sstar , m < \tilh , \cU_{m} \models \ga \}
\subseteq \{ m \mid m \in {\bf N}\sstar , r\sstar(m) \geq x \} . \]
Let us  define now
\[ \eta = \sumstar_{m \in \tilh , r\sstar(m) > x} \tilf(m) \]
and
\[ \rho = \sumstar_{m \in \tilh , r\sstar(m) = x} \tilf(m) . \]
For every formula {\gc} define:
\[ \lambda(\gc) = \sumstar_{m\in\tilh , r\sstar(m)=x , 
\, \cU_{m} \models \gc} \tilf(m) \]
Of course one always has \mbox{$Pr(\gc) \geq \lambda(\gc)$}.
Note that by Equation~\ref{eq:ratio2}, \mbox{$\eta / \rho$} is infinitesimal.
Also by item~\ref{itemstd} of the definition of the sequence 
\mbox{$\langle B_{n} \mid n \in {\bf N} \rangle$} if 
\mbox{$A_{\gc} \cap \{ m \mid m \in \tilh , r\sstar(m) = x \} \neq \emptyset$}
and if \mbox{$\gc = \ga_{j-1}, j \in {\bf N}$} then
\[ \lambda(\gc) \geq \rho \times \frac{1}{2^{j}}. \]
Now, assume \mbox{$\ab \in K$}.
Hence for every 
\mbox{$m \in \tilh$}, if \mbox{$\cU \models \neg \gb \wedge \ga$}, 
we must have \mbox{$r\sstar(m) > x$}.
Therefore \mbox{$Pr(\neg\gb\wedge\ga) \leq \eta$}.
Therefore:
\[ Pr(\neg\gb \mid \ga) = \frac{Pr(\neg\gb\wedge\ga)}{Pr(\ga)} \leq 
\frac{\eta}{\rho \times \frac{1}{2^{i}}} 
= 2^{i} \times \frac{\eta}{\rho} . \]
Therefore \mbox{$Pr(\neg\gb \mid \ga)$} is infinitesimal and by definition 
\mbox{$\ab \in K(\cM)$}.
If \mbox{$\ab \not\in K$}, then some \mbox{$m\in {\bf N}$}, 
\mbox{$r\sstar(m) = x$} satisfies \mbox{$\cU_{m} \models \neg\gb\wedge\ga$}.
But this $m$ satisfies \mbox{$m\in\tilh$}, so it is in our model.
Let \mbox{$j\in {\bf N}$} be such that \mbox{$\neg\gb\wedge\ga = \ga_{j-1}$}.
Since we clearly have: $Pr(\ga)\leq\rho + \eta$, we also have:
\[ Pr(\neg\gb\mid\ga) = \frac{Pr(\neg\gb\wedge\ga)}
{Pr(\ga)}\geq\frac{\frac{1}{2^{j}}
\times \rho} {\rho+\eta} \geq \frac{1}{2^{j+1}} \]
since obviously \mbox{$\rho \geq \eta$}.
So \mbox{$Pr(\neg\gb \mid \ga)$} is {\em not} infinitesimal
and \mbox{$\ab \not\in K(\cM)$}.
(end of proof of Claim~\ref{main-claim}) \QED
We have already noticed that \cM\ is a neat model. Claim~\ref{main-claim}
shows that it has the desired property.
(end of proof of Theorem~\ref{completeness-nonstd}) \QED

\end{document}